\documentclass{article}

\usepackage[main, final]{neurips_2026}

\usepackage[utf8]{inputenc} %
\usepackage[T1]{fontenc}    %
\usepackage{hyperref}       %
\usepackage{url}            %
\usepackage{booktabs}       %
\usepackage{amsfonts}       %
\usepackage{nicefrac}       %
\usepackage{microtype}      %
\usepackage{xcolor}         %

\usepackage{orcidlink}
\usepackage{multicol}
\usepackage{multirow}
\usepackage[accsupp]{axessibility} %
\usepackage{amssymb}%
\usepackage{pifont}%
\usepackage{appendix}
\usepackage{graphicx}
\usepackage{enumitem, romannum, titletoc}
\newcommand{\cmark}{\ding{51}}%
\newcommand{\xmark}{\ding{55}}%
\usepackage[table]{xcolor}
\newcommand{\methodname}{TimeTraveler} %

\definecolor{softblue}{RGB}{220,235,252}

\usepackage[most]{tcolorbox} 

\usepackage{placeins}

\title{TimeTraveler: Temporal Strategy Planning with \\Time Dictionary for Streaming Video Understanding}

\author{
Minkuk Kim$^{1}$~~~~
Junho Kim$^{2}$\addtocounter{footnote}{1}%
\thanks{Corresponding authors.}~~~~
Seong Tae Kim$^{1}$\footnotemark[2]
\\[2mm]
$^{1}$Kyung Hee University~~~~
$^{2}$University of Illinois Urbana-Champaign
\\
}

\begin{document}

\pagenumbering{arabic}
\maketitle

\begin{abstract}
Streaming video understanding (SVU) requires models to answer user queries over continuously evolving video streams. Unlike offline video QA, SVU must handle queries whose evidence may lie in the past, appear in the current scene, or emerge only after future observations. Existing streaming video systems often rely on compact visual memories or fixed evidence-access schemes, which can lose fine-grained details and fail to adapt the answering process to the temporal intent of each query. In this paper, we propose TimeTraveler, a streaming VQA framework that performs temporal strategy planning with a Time Dictionary. Inspired by the need to search across different temporal regions of a stream, TimeTraveler stores each observed moment as a timestamp-indexed structured caption and uses this dictionary as an explicit source of queryable evidence. Given a query, TimeTraveler determines whether to recall past evidence, read the present context, or wait for future observations, and then applies a strategy-specific evidence acquisition process. Comparative experiments on streaming video QA benchmarks show that TimeTraveler improves streaming question answering by preserving fine-grained temporal evidence and selecting the appropriate temporal strategy for each query. The project page is available at \url{https://geppa.github.io/TimeTraveler}.

\end{abstract}

\section{Introduction}

The rapid progress of Large Vision-Language Models (LVLMs) has expanded the scope of video understanding~\citep{bai2025qwen2,li2024llava,li2025llama,vicuna,wang2025internvl3,zhang2025videollama,maaz2024video}, shifting the field from task-specific recognition toward open-ended visual reasoning~\citep{wang2025videotree,feng2025video,li2025videochat}, temporal localization~\citep{liu2024end,ye2025re}, and interactive question answering~\citep{ren2024timechat,jin2024chat}. 
Despite this progress, most video-LLMs~\cite{liu2025nvila,Qwen3-VL} are still designed for an offline assumption: a holistic video is given in advance, and the model answers after processing the entire input. 
This assumption breaks down in streaming scenarios, where visual observations arrive causally, the stream may be unbounded, and user queries can be issued at arbitrary moments. 
In such settings, the required evidence is structurally heterogeneous. The system must not only support rich perception of incoming visual information, but also preserve past events for later access and remain responsive to events that can only be confirmed through future observation. These requirements make streaming video understanding (SVU) particularly challenging.

Recent studies on SVU fall into two main directions. 
One develops memory-centric mechanisms~\citep{zhang2025flash,yang2025streammem,xiong2025streaming} that compress and retain stream observations (\textit{e.g.,} prototypes, memory tokens, hierarchical summaries, or compressed KV caches) to preserve past information under limited budgets. 
The other branch~\citep{wang2025streambridge,chen2024videollm,zhang2026querystream,qian2025dispider} improves online interaction by enabling real-time response scheduling, proactive activation, or streaming adaptation of offline video MLLMs, allowing the system to respond at more appropriate moments as the stream unfolds. 

\begin{figure*}[t]
    \centering
    \includegraphics[width=0.92\linewidth]{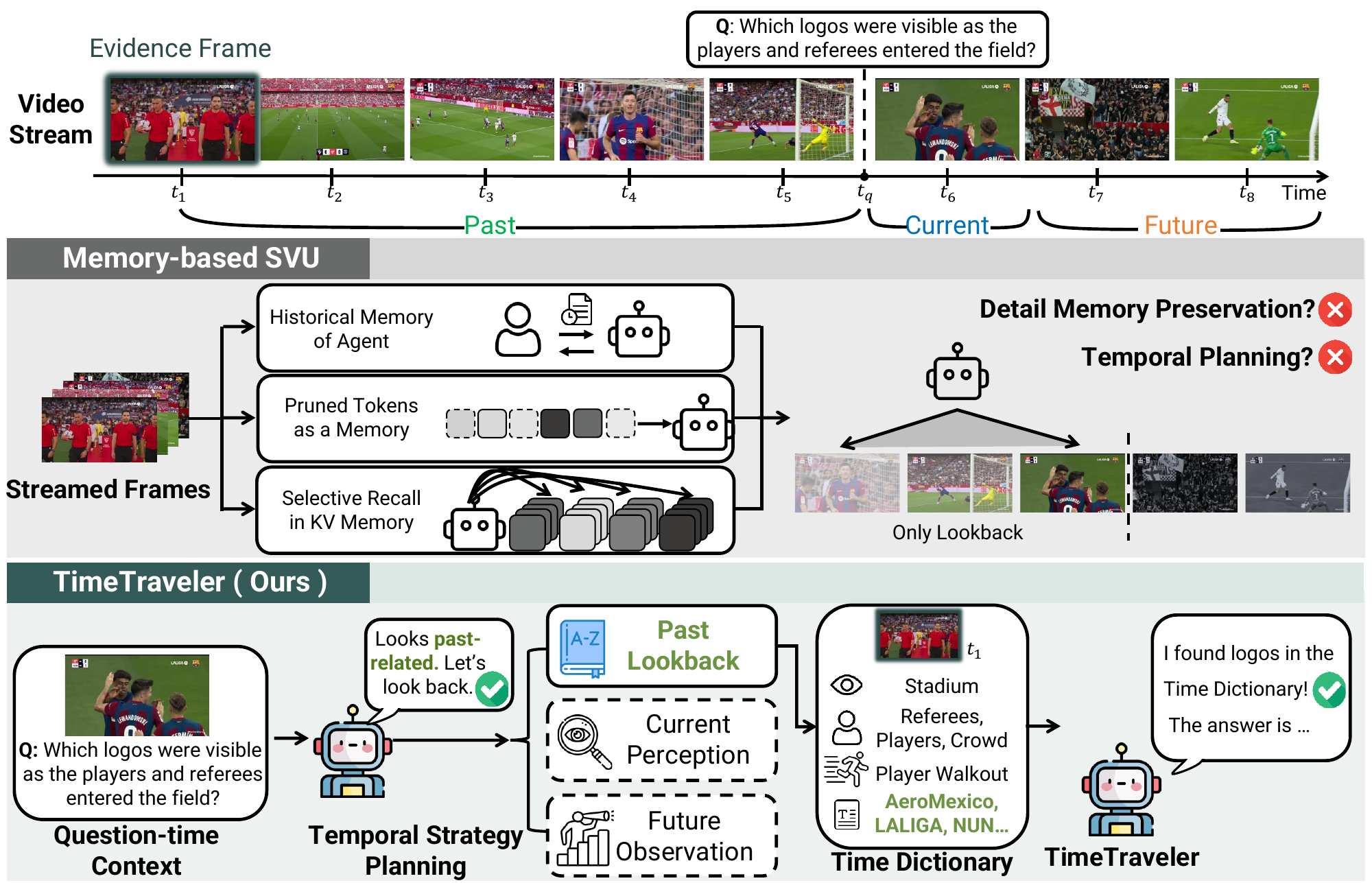}
    \caption{
    \textbf{Conceptual comparison between prior memory-based SVU and \methodname{}.}
    Prior methods preserve streaming history as agentic memory~\cite{xiong2025streaming}, pruned tokens~\cite{zhang2026querystream}, or KV-memory recall~\cite{yang2025streamagent}, but rely on a single evidence-access path and do not ensure that retained evidence remains query-recoverable. In contrast, \methodname{} plans whether each query should be grounded in the past, current scene, or future, and accesses a structured \textit{Time Dictionary} accordingly.
    }
    \vspace{-0.3cm}
\label{fig:teaser}
\end{figure*}

Specifically, memory-based methods preserve evidence by converting the video stream into historical memory~\citep{xiong2025streaming}, caching visual information as KV memory~\citep{yang2025streamagent,yang2025streammem}, or selectively storing query-relevant information via token pruning~\citep{zhang2026querystream,yao2025timechat}. To make unfolding streams tractable, these methods often encode streamed observations into latent tokens, visual prototypes, or compressed representations for stable and efficient retention. Although these approaches are effective for streaming memory, such representations can obscure fine-grained cues including small text, object attributes, spatial relations, or transient actions, making them difficult to recover at question time, as shown in Fig.~\ref{fig:teaser}. Since future queries are unknown during observation, cues that seem incidental when stored may later become answer-critical. Thus, beyond retaining a compressed history, SVU requires an explicit, fine-grained, and queryable memory representation that keeps such details recoverable at question time.

Moreover, preserving recoverable evidence alone does not determine how that evidence should be accessed for an arbitrary user query. Prior approaches~\citep{xiong2025streaming,yang2025streammem,yao2025timechat} largely overlook when the evidence for answering a query should be sought, without explicitly modeling the temporal structure of user queries. We argue that SVU is fundamentally a temporal grounding problem, where each query implicitly specifies a temporal scope $\{\mathrm{past}, \mathrm{current}, \mathrm{future}\}$. Accordingly, current queries rely on local evidence, past queries require retrieval from accumulated observations, and future queries must be monitored until the relevant event occurs.

To bridge this gap, in this paper, we propose \methodname{}, a SVU framework that stores incoming observations as an explicit, fine-grained, and queryable memory representation, and accesses the stored evidence according to the temporal scope for the given user query. Specifically, \methodname{} is built on two components: (\lowercase\expandafter{\romannumeral1}) Time Dictionary, which stores the stream as lightweight structured per-frame captions that decompose visual information within each frame into multiple semantic fields. Each timestamp is associated with explicit textual evidence covering scenes, objects, actions, text, spatial relations, and contextual details, making the memory both directly queryable and efficient to maintain in real-time streaming settings, and (\lowercase\expandafter{\romannumeral2}) Temporal Strategy Planner for deciding whether a query should be answered from past memory, the current scene, or future observation. The model then follows the selected temporal strategy: for past queries, it performs memory lookback; for current queries, it refers to structured perception of the current scene; and for future queries, it carries out structured observation until the required evidence becomes available, all within a unified SVU system.

Through experiments on two streaming benchmarks~\cite{lin2026streamingbench,niu2025ovo}, we show that \methodname{} consistently outperforms various baselines. More importantly, the results support our central claim that SVU depends not only on retaining fine-grained history (Table~\ref{tab:text_vs_visual}), but also on accessing the right evidence according to the temporal intent of each query (Table~\ref{tab:routing_effect}). By combining temporal strategy planning with a structured Time Dictionary, \methodname{} improves evidence access across past, current, and future contexts, enabling more effective question answering in video streams.

Our contributions are summarized as follows:
\begin{itemize}[leftmargin=1.2em, itemsep=0.2em, topsep=0.2em]

    \item We propose \methodname{}, a streaming video QA framework that stores fine-grained evidence in a lightweight Time Dictionary and retrieves it through query-aware Temporal Strategy Planning. Unlike compressed visual memory, our design preserves query-recoverable evidence over long streaming horizons while supporting efficient updates.

    \item We formulate SVU as a temporal grounding problem and identify temporal routing as a missing inductive bias. We introduce Temporal Strategy Planning, a query-aware mechanism that selects how evidence should be acquired from past, current, or future contexts.

    \item We validate \methodname{} across multiple backbones, streaming video benchmarks, and efficiency metrics, and provide mechanistic analyses showing that temporal strategy planning and the Time Dictionary improve evidence selection and streaming question answering quality.
\end{itemize}

\section{Related Work}

\subsection{Large Vision-Language Models}

Early LVLMs~\cite{liu2023visual, dai2023instructblip, liu2023improved} show that visual instruction tuning can connect pretrained language backbones~\citep{touvron2023llama, vicuna} with visual encoders for general-purpose multimodal conversation. Subsequent models improve this paradigm by scaling model and data, refining visual tokenization, and strengthening vision-language alignment~\citep{li2024llava, zhu2025internvl3, wang2025internvl3, bai2025qwen2}. Video-oriented LVLMs~\citep{lin2023video, li2024mvbench, kim2024salova, zhang2025videollama, li2025llama} further extend these capabilities to temporal inputs by treating videos as sequences of images and introducing video-specific connectors, temporal aggregation modules, or video instruction-tuning pipelines. Caption-based approaches instead convert clips into text and reason over it with an LLM~\citep{zhang2024llovi,pang2025mrvideo}, and retrieval over external memory further aids video and visual QA~\citep{kim2024cm2,kim2025hicm2,lim2024retrieval}. Despite strong progress on video QA and captioning, most of these methods are designed for offline or fixed-context inference, where the video is fully available before answering or has already been compressed into a finite context window. This assumption is restrictive for SVU, where frames arrive causally and a query may require evidence from past history, the current scene, or future observations.

\subsection{Streaming Video Understanding}

Streaming video understanding studies how multimodal models process continuously arriving visual evidence before the full video is available. A major line of work addresses the retention problem: how to maintain useful history from an unbounded stream under bounded computation and memory. StreamChat builds hierarchical memory for long-stream and multi-round interaction~\citep{liu2024streamchat}, Flash-VStream maintains compact memories for real-time long-video response~\citep{zhang2025flash}, and StreamMem reduces streaming memory cost through visual cache compression~\citep{yang2025streammem}. Related memory-based or prototype-based systems summarize incoming observations into compact token memories, learned prototypes, or recurrent retained states~\citep{song2024moviechat,xiong2025streaming}, and InfiniPot-V~\citep{kim2025infinipot} enforces a length-independent cap on the visual KV cache through training-free compression during streaming. These methods make long-horizon processing practical by compressing accumulated context or reducing redundant visual tokens. In contrast, our framework keeps the retained history as explicit structured caption records: each sampled frame is decomposed into semantic fields such as actions, people, objects, scene context, text, and spatial relations, so the memory remains more structured and query-recoverable at question time rather than being collapsed into a compact visual state.

Another line of work focuses on online interaction and proactive response, asking when a streaming model should produce an output. VideoLLM-Online and related systems adapt Video-LLMs to interleaved video-text streams~\citep{chen2024videollm}, while Dispider decomposes real-time interaction into perception, decision, and reaction modules~\citep{qian2025dispider}. StreamBridge converts offline Video-LLMs into proactive streaming assistants with a memory buffer and lightweight activation module~\citep{wang2025streambridge}, and QueryStream introduces query-aware pruning with relevance-triggered active response scheduling~\citep{zhang2026querystream}. These approaches improve response timing and streaming efficiency, but they primarily decide whether or when the answer model should be activated, rather than explicitly modeling the temporal grounding decision implied by the user's query. In contrast, our work makes this intermediate decision explicit: before assembling evidence, the system first determines whether the query should be grounded in past memory, the current local context, or future observations, and then applies the corresponding evidence strategy over structured memory. Unlike tool selection in LLM agents~\citep{schick2023toolformer,shen2023hugginggpt} or query-aware selection over already observed content~\citep{zhang2026querystream,yao2025timechat,kim2026request}, our planner predicts where in time the evidence lies, including evidence that has not yet appeared.

\begin{figure*}[t]
    \centering
    \includegraphics[width=0.95\linewidth]{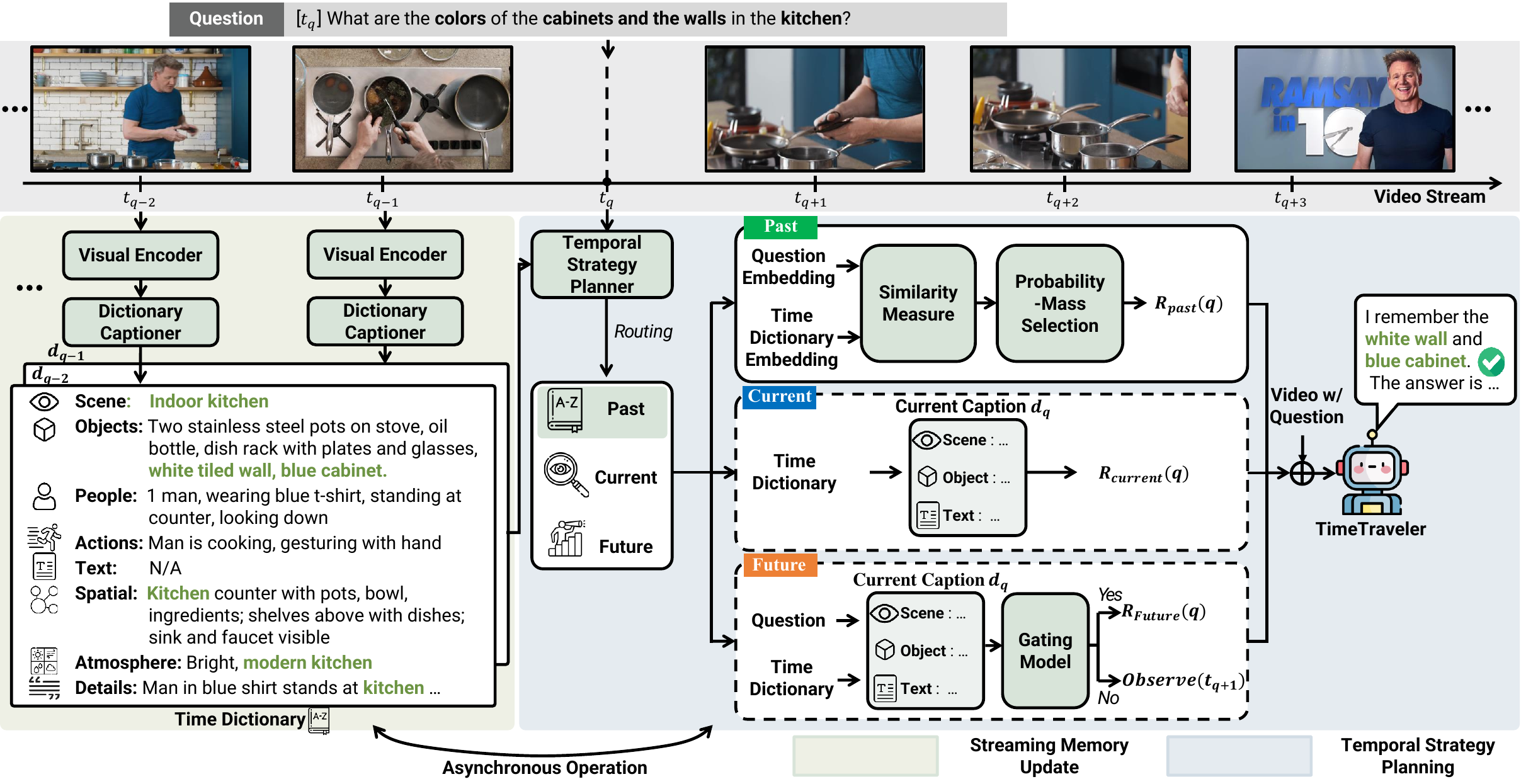}
    \vspace{-0.2cm}
    \caption{
    \textbf{Overview of \methodname{}.} Incoming video frames are encoded into structured captions and stored in a timestamp-indexed Time Dictionary. Given a query, a Temporal Strategy Planner acquires evidence following temporal intent: past queries retrieve relevant captions through similarity-based lookback, current queries access local captions, and future queries continue observation until evidence appears. Then, the retrieved evidence is used to produce the final answer.
    }
    \vspace{-0.3cm}
\label{fig:overall}
\end{figure*}

\section{Proposed Method}
\label{sec:method}

\paragraph{Overview.} We propose \methodname{}, a streaming framework for query-conditioned evidence acquisition, as shown in Fig.~\ref{fig:overall}. 
Our framework separates what to retain from the stream and how to access it for each query. Incoming frames are converted into structured caption records and stored in a causal Time Dictionary, together with sentence embeddings used only as lightweight retrieval handles.
When a query arrives, a Temporal Strategy Planner selects one of three temporal strategies: past, current, or future. These respectively retrieve prior evidence with recency-aware lookback, read the query-time caption, or continue observing until a response condition or horizon is reached.

\subsection{Time Dictionary for Structured Evidence Retention}
\label{sec:method:memory}

The goal of the Time Dictionary is to retain queryable evidence from a causal video stream without collapsing the history only into a visual prototype or recurrent hidden state. Rather than compressing observations into latent visual states, we represent the stream as a temporally indexed structured representation that preserves fine-grained, query-recoverable evidence.
Let $f_i$ denote the $i$-th sampled frame and $\tau_i$ its timestamp. 
At each observed moment, sampled at 1 fps in our streaming pipeline, a frozen captioner $\Phi_{\mathrm{cap}}$ processes the frame once following eight fields: scene, objects, people, actions, text, spatial relations, atmosphere, and details. 
This produces a bounded-length structured caption $c_i = \Phi_{\mathrm{cap}}(f_i)$, written in a consistent schema of the form:
\begin{equation}
\small
\begin{aligned}
    c_i = \{&\texttt{Scene}: s_i^{\mathrm{scene}},\; \texttt{Objects}: s_i^{\mathrm{obj}},\; \texttt{People}: s_i^{\mathrm{people}},\; \texttt{Actions}: s_i^{\mathrm{act}},\\
    &\texttt{Text}: s_i^{\mathrm{text}},\; \texttt{Spatial}: s_i^{\mathrm{spatial}},\; \texttt{Atmosphere}: s_i^{\mathrm{atm}},\; \texttt{Details}: s_i^{\mathrm{det}}\}.
\end{aligned}
\normalsize
\end{equation}
Here, \texttt{Scene} summarizes locational and environmental context such as indoor or outdoor setting; \texttt{Objects} lists salient entities together with attributes such as color, size, and state; \texttt{People} describes count, appearance, clothing, and expression; \texttt{Actions} captures ongoing events and interactions among people and objects; \texttt{Text} records any visible signs, labels, subtitles, or logos; \texttt{Spatial} specifies layout and relative positions of key elements; \texttt{Atmosphere} summarizes lighting, color tone, time of day, weather, or mood; and \texttt{Details} provides a short coherent paragraph that captures contextual cues not fully covered by the other fields. 
The schema encourages the captioner to explicitly expose fine-grained cues that may later become relevant to an arbitrary query. 

For efficient lookback, the system also computes an auxiliary sentence embedding for each caption. 
A frozen text encoder $\Phi_{\mathrm{emb}}$ encodes the full caption as one textual sequence and L2-normalizes it to $e_i$.
For each sampled frame $i$, we create a dictionary entry $d_i = (i, \tau_i, c_i, e_i)$, and the Time Dictionary is then updated causally by appending the new entry, $\mathcal{D}_{i} = \mathcal{D}_{i-1} \Vert d_i$, with $\mathcal{D}_0 = ()$.

Thus, when a query arrives at frame index $i_q$, the system can only access $\mathcal{D}_{i_q}$, ensuring that no future frame is used for past or current answering. 
The Time Dictionary preserves streamed observations as explicit timestamped text records, allowing fine-grained evidence to remain available over long time horizons without being diluted as the stream unfolds. The auxiliary embeddings serve only as lightweight handles for efficient lookback, not as the primary memory content. For memory-efficient retrieval, we maintain these embeddings in a finite bin queue covering roughly the most recent 600 seconds of the stream, while the text-based Time Dictionary remains the persistent evidence store. This horizon bounds index growth on potentially unbounded streams and can be extended at negligible cost (Appendix~\ref{app:horizon}). This timestamp-indexed structure makes the accumulated stream addressable by different temporal strategies rather than forcing all queries to share a single compact memory state.

\subsection{Query-Conditioned Temporal Strategy Planning}
\label{sec:method:policy}

We then formulate temporal strategy selection as a query-conditioned policy over a discrete temporal action space $\{\mathrm{past}, \mathrm{current}, \mathrm{future}\}$. Rather than predicting an answer directly, the policy determines how evidence should be acquired from the streaming setup. For example, a question such as ``\textit{What did the person pick up earlier?}'' requires retrieving evidence from past memory, whereas ``\textit{What is the person doing now?}'' is answered from the current scene. 
A future-facing instruction such as ``\textit{Tell me when the car stops.}'' should not be answered immediately, but should instead be monitored until the requested condition becomes true or the observation horizon is reached. Temporal Strategy Planning captures these distinctions by assigning each query to the strategy that matches its temporal scope.

\paragraph{Temporal strategy planner.}
Given a query $q$ issued at timestamp $\tau_q$ and latest observed frame index $i_q$, the planner serializes $q$ together with a short recent-caption context and prompts a frozen lightweight instruction model to output exactly one temporal strategy $\pi(q)\in\{\mathrm{past},\mathrm{current},\mathrm{future}\}$.
The prompt instructs the model to choose \texttt{future} when the query asks the system to wait for a condition that has not yet occurred, \texttt{past} when the answer requires information outside the recent context, and \texttt{current} otherwise. The role of the planner is not to answer the question, but to decide how evidence should be sought before answer generation.

\paragraph{Past strategy: online probabilistic lookback.}
For past-directed queries, the system performs online probabilistic lookback over the observed Time Dictionary $\mathcal{D}_{i_q}$. 
Rather than selecting a fixed top-$K$ set, the past strategy first forms a similarity-based candidate pool and then adaptively selects evidence using a recency-aware softmax distribution.

We first restrict the search to causal entries whose timestamps do not exceed the query time:
\begin{equation}
    \mathcal{I}_q = \{ i \mid d_i \in \mathcal{D}_{i_q},\ \tau_i \le \tau_q \}.
\end{equation}
From this causal set, we compute a semantic similarity score $u_i=\cos(\phi(q), e_i)$ for each $i\in\mathcal{I}_q$, where $\phi(q)$ denotes the normalized query embedding.
We keep the top-$N$ entries by $u_i$ as a candidate index set $\mathcal{C}_N(q) \subseteq \mathcal{I}_q$.
In our implementation, we use $N=32$.

For each candidate entry, we form a recency-aware selection logit:
\begin{equation}
    \ell_i
    =
    u_i
    -
    \frac{\tau_q-\tau_i}{\tau_r},
    \qquad i \in \mathcal{C}_N(q),
\label{eq:past_logit}
\end{equation}
where $\tau_r$ controls how strongly older evidence is penalized.
The logits are converted into a probability distribution with temperature $T_{\mathrm{past}}$:
\begin{equation}
    p_i
    =
    \frac{\exp(\ell_i / T_{\mathrm{past}})}
    {\sum_{j \in \mathcal{C}_N(q)} \exp(\ell_j / T_{\mathrm{past}})},
    \qquad i \in \mathcal{C}_N(q).
\label{eq:past_prob}
\end{equation}

We then sort candidate indices in descending order of $p_i$, denoted by $(1),(2),\ldots,(|\mathcal{C}_N(q)|)$, and select the smallest prefix whose cumulative probability mass exceeds a threshold $t_p$:
\begin{equation}
    K^\star
    =
    \min
    \left\{
    k :
    \sum_{m=1}^{k} p_{(m)} \ge t_p
    \right\}.
\label{eq:past_nucleus_k}
\end{equation}
We define the past evidence set as $\mathcal{R}_{\mathrm{past}}(q)=\{ d_{(m)} \mid 1 \le m \le K^\star \}$.
This cumulative-mass cutoff makes the number of selected past entries adaptive: concentrated distributions yield a small evidence set, while diffuse distributions retain more candidates. 
Importantly, embeddings are used only to construct the candidate pool and selection distribution; the evidence passed to the answer model is the stored caption text. 
This strategy supports non-local lookback when the answer depends on earlier observations that may no longer be visible.

\paragraph{Current strategy: query-time caption.}
For current-directed queries, global memory lookback is unnecessary and can introduce distractors from semantically similar but temporally irrelevant past events. 
The current strategy therefore reads the caption record at the query time as $\mathcal{R}_{\mathrm{current}}(q)=\{d_{i_q}\}$.
This provides direct access to the dictionary caption record at the query time and avoids mixing current evidence with temporally irrelevant past entries.

\paragraph{Future strategy: observe-or-respond gating.}
For future-directed queries, the system enters a deferred observation mode. 
Rather than invoking the answer model immediately, it periodically checks whether later evidence satisfies the query condition. 
Let $W \in \mathbb{N}_{>0}$ denote the evidence window size, i.e., the number of most recent dictionary entries passed to the gate. 
A frozen lightweight gate model $g$, prompted with a fixed yes/no instruction, takes the query and the window $\mathcal{D}_{i-W+1:i}$ and predicts whether the condition has been satisfied:
\begin{equation}
    b_i
    =
    g(q, \mathcal{D}_{i-W+1:i})
    \in \{0,1\},
    \qquad i>i_q.
\end{equation}
The system polls the gate at intervals of $\Delta_{\mathrm{poll}} \in \mathbb{N}_{>0}$ frames after $i_q$, and continues polling up to a maximum observation horizon $T_{\max}$ (in seconds) measured from the query time $\tau_q$. 
The first polling index whose gate output is positive is selected:
\begin{equation}
    i^\star
    =
    \min
    \{i>i_q : i \equiv i_q \!\!\!\pmod{\Delta_{\mathrm{poll}}},
    \ \tau_i-\tau_q \le T_{\max},
    \ b_i=1\}.
\label{eq:future_gate}
\end{equation}
If no polling index satisfies the condition before the maximum horizon $T_{\max}$, the system falls back to the latest available evidence under the observation budget. 
If no future evidence is available, it falls back to the caption record at the query time. 
This strategy enables delayed response without repeatedly calling the full answer model: the gate is evaluated during observation, and the answer model is invoked only once after the gate fires or the horizon is reached.

\paragraph{Evidence assembly.}
Let $\mathcal{R}_{\mathrm{past}}(q)$, $\mathcal{R}_{\mathrm{current}}(q)$, and $\mathcal{R}_{\mathrm{future}}(q)$ denote the evidence returned by the three strategies. 
The final evidence sequence is selected according to the planned temporal strategy:
\begin{equation}
\mathcal{R}(q)=
\begin{cases}
\mathcal{R}_{\mathrm{past}}(q), & \pi(q)=\mathrm{past},\\
\mathcal{R}_{\mathrm{current}}(q), & \pi(q)=\mathrm{current},\\
\mathcal{R}_{\mathrm{future}}(q), & \pi(q)=\mathrm{future}.
\end{cases}
\label{eq:evidence_assembly}
\end{equation}
The selected entries are serialized with their timestamps and captions, and are then passed to a frozen long-context answer model $\mathrm{LM}_{\mathrm{QA}}$, which generates the final response. 

\subsection{Streaming Inference}
\label{sec:method:inference}

At inference time, the system runs an observe-and-respond loop. 
The ingest process continuously applies $\Phi_{\mathrm{cap}}$ and $\Phi_{\mathrm{emb}}$ to incoming frames and appends the resulting entries to $\mathcal{D}$. Streaming updates are efficient because each sampled frame produces one bounded caption record and one embedding append, while future queries require only lightweight periodic gating before a single final answer-generation step.
When a query arrives, the dispatch process prompts the Temporal Strategy Planner once. 
Past and current queries invoke $\mathrm{LM}_{\mathrm{QA}}$ immediately using evidence available at $i_q$. 
Future queries create a polling state and evaluate the prompt-based gate only at newly reached polling indices while the ingest process continues updating the Time Dictionary. 

Once the gate fires, or the maximum horizon is reached, the system invokes $\mathrm{LM}_{\mathrm{QA}}$ exactly once with the selected future evidence. At QA time, the answer model takes as input the query $q$, the video observations available at decision time, and the temporal evidence assembled by our method, $\mathcal{R}(q)$, and produces the final VQA answer, $a = \mathrm{LM}_{\mathrm{QA}}\bigl(v, q, \mathcal{R}(q)\bigr)$, 
where $v$ denotes the video frames available when the system answers the query and $a$ is the generated answer.

\vspace{-1mm}
\section{Experiments}
\label{sec:experiments}
\vspace{-1mm}
\subsection{Experimental Setting}

\noindent\textbf{Benchmarks and Metrics.}
We evaluate \methodname{} on two complementary streaming video understanding benchmarks: OVO-Bench~\citep{niu2025ovo} and StreamingBench~\citep{lin2026streamingbench}. OVO-Bench measures three challenging capabilities in online video understanding: real-time visual perception, backward tracing, and forward active responding. Following the benchmark protocol, we report category-wise accuracies together with the macro average across the three taxonomies as the overall score. StreamingBench evaluates streaming comprehension over 18 tasks spanning real-time visual understanding, omni-source understanding, and contextual understanding. %

\noindent\textbf{Implementation Details.}
\methodname{} builds on the Qwen3 family~\citep{Qwen3-VL}: it uses a frozen 8-bit-quantized Qwen3-VL-2B as both the visual captioner and the planner/gate model, Qwen3-Embedding-0.6B as the text encoder, and Qwen3-VL-8B as the frozen answer model; we also evaluate with InternVL3.5-8B~\citep{wang2025internvl3}. Incoming frames are sampled at 1 FPS and converted into structured caption records with eight semantic fields, while one auxiliary embedding per caption is maintained only as a retrieval handle in a finite cache covering roughly the most recent 600 seconds. For past-directed queries, we use a candidate pool size of $N=32$ and set $\tau_r=150$, $T_{\mathrm{past}}=0.02$, and $t_p=0.70$; for future-directed queries, we set $\Delta_{\mathrm{poll}}=1$ frame, $T_{\max}=30$ s, and $W=1$. The answer model receives the 1-fps frames observed up to the decision time, capped at 512 frames for Qwen3-VL-8B and 128 frames for InternVL3.5-8B due to context limits; all other settings remain identical across backbones. All components are used in frozen form and only prompt-based planning is applied at inference time. Experiments are run on a single NVIDIA RTX PRO 6000 Blackwell GPU.

\begin{table*}[t]
\centering
\caption{\textbf{Evaluation results on OVO-Bench.} Overall accuracy is calculated as the macro average across taxonomies. Best results among open-source models are \textbf{boldfaced} and runner-up results are \underline{underlined}. Proprietary model results are shown with gray background for reference.}
\label{tab:bench_ovo}
\vspace{2mm}
\resizebox{\textwidth}{!}{%
\begin{tabular}{l c|ccccccc|cccc|cccc|c}
\toprule
Method  & \#Frames
& \multicolumn{7}{c|}{Real-Time Visual Perception}
& \multicolumn{4}{c|}{Backward Tracing}
& \multicolumn{4}{c|}{Forward Act. Responding}
& Overall \\
\cmidrule(lr){3-9} \cmidrule(lr){10-13} \cmidrule(lr){14-17} \cmidrule(lr){18-18}
& 
& OCR & ACR & ATR & STU & FPD & OJR & Avg.
& EPM & ASI & HLD & Avg.
& REC & SSR & CRR & Avg.
& Avg. \\
\midrule
\rowcolor{gray!35}
\multicolumn{18}{c}{\textbf{Human}} \\
\rowcolor{gray!15}
Human & -
& 93.96 & 92.57 & 94.83 & 92.70 & 91.09 & 94.02 & 93.20
& 92.59 & 93.02 & 91.37 & 92.33
& 95.48 & 89.67 & 93.56 & 92.90
& 92.81 \\
\midrule
\rowcolor{gray!35}
\multicolumn{18}{c}{\textbf{Proprietary Models (Offline)}} \\
\rowcolor{gray!15}
Gemini 1.5 Pro~\citep{team2024gemini} & 1 fps
& 85.91 & 66.97 & 79.31 & 58.43 & 63.37 & 61.96 & 69.32
& 58.59 & 76.35 & 52.64 & 62.54
& 35.53 & 74.24 & 61.67 & 57.15
& 63.00 \\
\rowcolor{gray!15}
GPT-4o~\citep{hurst2024gpt} & 64
& 69.80 & 64.22 & 71.55 & 51.12 & 70.30 & 59.78 & 64.46
& 57.91 & 75.68 & 48.66 & 60.75
& 27.58 & 73.21 & 59.40 & 53.40
& 59.54 \\
\midrule
\multicolumn{18}{c}{\textbf{Open-Source Offline Models}} \\
LLaVA-Video-7B~\citep{lin2023video} & 64
& 69.13 & 58.72 & 68.83 & 49.44 & \underline{74.26} & 59.78 & 63.52
& \underline{56.23} & 57.43 & 7.53 & 40.40
& \underline{34.10} & \textbf{69.95} & \underline{60.42} & \textbf{54.82}
& 52.91 \\
LLaVA-OV-7B~\citep{li2024llava} & 64
& 66.44 & 57.80 & 73.28 & 53.37 & 71.29 & 61.96 & 64.02
& 54.21 & 55.41 & 21.51 & 43.71
& 25.64 & 67.09 & 58.75 & 50.50
& 52.74 \\
LLaVA-N-Video-7B~\citep{li2024llavaNinterleave} & 64
& 69.80 & 59.60 & 66.40 & 50.60 & 72.30 & 61.40 & 63.30
& 51.20 & 64.20 & 9.70 & 41.70
& \underline{34.10} & 67.60 & \textbf{60.80} & \underline{54.20}
& \underline{53.10} \\
Qwen2-VL-7B~\citep{yang2024qwen2}  & 64
& 60.40 & 50.46 & 56.03 & 47.19 & 66.34 & 55.43 & 55.98
& 47.81 & 35.48 & \textbf{56.08} & \underline{46.46}
& 31.66 & 65.82 & 48.75 & 48.74
& 50.39 \\
InternVL-V2-8B~\citep{chen2024far} & 64
& 67.11 & 60.55 & 63.79 & 46.07 & 68.32 & 56.52 & 60.39
& 48.15 & 57.43 & 24.73 & 43.44
& 26.50 & 59.14 & 54.14 & 46.60
& 50.15 \\
LongVU-7B~\citep{shen2024longvu} & 1 fps
& 55.70 & 49.50 & 59.50 & 48.30 & 68.30 & 63.00 & 57.40
& 43.10 & 66.20 & 9.10 & 39.50
& 16.60 & \underline{69.00} & 60.00 & 48.50
& 48.50 \\
\midrule
\multicolumn{18}{c}{\textbf{Open-Source Online Models}} \\
Flash-VStream-7B~\citep{zhang2025flash} & 1 fps
& 24.16 & 29.36 & 28.45 & 33.71 & 25.74 & 28.80 & 28.37
& 39.06 & 37.16 & 5.91 & 27.38
& 8.02 & 67.25 & 60.00 & 45.09
& 33.61 \\
VideoLLM-Online-8B~\citep{chen2024videollm} & 2 fps
& 8.05 & 23.85 & 12.07 & 14.04 & 45.54 & 21.20 & 20.79
& 22.22 & 18.80 & 12.18 & 17.73
& - & - & - & -
& - \\
VideoLLM-EyeWO~\citep{zhang2025eyes} & 1 fps
& 24.16 & 27.52 & 31.89 & 32.58 & 44.55 & 35.87 & 32.76
& 39.06 & 38.51 & 6.45 & 28.00
& - & - & - & -
& - \\
Dispider-7B~\citep{qian2025dispider} & 1 fps
& 57.72 & 49.54 & 62.07 & 44.94 & 61.39 & 51.63 & 54.55
& 48.48 & 55.41 & 4.30 & 36.06
& 18.05 & 37.36 & 48.75 & 34.72
& 41.78 \\
TimeChat-Online-7B~\citep{yao2025timechat} & 1 fps
& 69.80 & 48.60 & 64.70 & 44.90 & 68.30 & 55.40 & 58.60
& 53.90 & 62.80 & 9.10 & 42.00
& 32.50 & 36.50 & 40.00 & 36.40
& 45.60 \\
StreamAgent-7B~\citep{yang2025streamagent}  & 1 fps
& 71.20 & 53.20 & 63.60 & 53.90 & 67.30 & 58.70 & 61.30
& 54.80 & 58.10 & \underline{25.80} & 41.70
& \textbf{35.90} & 48.40 & 52.00 & 45.40
& 49.40 \\
QueryStream-7B~\citep{zhang2026querystream}  & 1 fps
& 74.50 & 47.70 & 70.70 & 46.60 & 71.30 & 57.60 & 61.40
& 54.20 & 63.50 & 8.60 & 42.10
& 33.20 & 43.10 & 40.80 & 39.03
& 47.51 \\
\midrule
InternVL3.5-8B~\citep{wang2025internvl3} & 1fps
& 79.87 & 65.14 & 69.83 & \textbf{61.80} & 70.30 & 65.22 & 68.69
& 33.00 & 59.46 & 4.30 & 32.25
& 32.66 & 60.10 & 35.00 & 42.59
& 47.84 \\

\rowcolor{softblue}
\quad + \methodname & 1fps
& \textbf{91.95} & \underline{70.64} & \underline{75.00} & \textbf{61.80} & \textbf{75.25} & \textbf{77.17} & \textbf{75.30}
& 33.67 & 60.81 & 9.68 & 34.72
& 22.21 & 50.24 & 38.75 & 37.06
& 49.03 \\
Qwen3-VL-8B~\citep{Qwen3-VL}  & 1fps
& 75.17 & 60.55 & 66.38 & 53.93 & 66.34 & 57.07 & 63.24
& 52.53 & \underline{67.57} & 10.22 & 43.44
& 33.09 & 65.50 & 46.67 & 48.42
& 51.70 \\
\rowcolor{softblue}
\quad + \methodname & 1fps
& \underline{87.25} & \textbf{73.39} & \textbf{76.72} & \underline{59.55} & 68.32 & \underline{74.46} & \underline{73.28}
& \textbf{59.26} & \textbf{69.59} & 25.27 & \textbf{51.37}
& 32.23 & 67.25 & 48.75 & 49.41
& \textbf{58.02} \\
\bottomrule
\end{tabular}%
}
\vspace{-4mm}
\end{table*}

\begin{table*}[t]
\centering
\setlength{\tabcolsep}{3pt}
\caption{\textbf{Performance comparison on StreamingBench.} Best results among open-source models are \textbf{boldfaced} and runner-up results are \underline{underlined}.}
\vspace{2mm}

\resizebox{\textwidth}{!}{%
\begin{tabular}{lc|ccccccccccc|ccccc|ccccc|c}
\toprule
Model & \#Frames
& \multicolumn{11}{c|}{Real-Time Visual Understanding}
& \multicolumn{5}{c|}{Omni-Source Understanding}
& \multicolumn{5}{c|}{Contextual Understanding}
& Overall \\
\cmidrule(lr){3-13} \cmidrule(lr){14-18} \cmidrule(lr){19-23}
&
& OP & CR & CS & ATP & EU & TR & PR & SU & ACP & CT & All
& ER & SCU & SD & MA & All
& ACU & MCU & SQA & PO & All
& \\
\midrule

\rowcolor{gray!30}
\multicolumn{24}{c}{\textbf{Human}} \\
\rowcolor{gray!15}
Human$^{\ddag}$  & -
& 89.47 & 92.00 & 93.60 & 91.47 & 95.65 & 92.52 & 88.00 & 88.75 & 89.74 & 91.30 & 91.46
& 88.00 & 88.24 & 93.60 & 90.27 & 90.26
& 88.80 & 90.40 & 95.00 & 100.00 & 93.55
& 91.66 \\
\midrule
\rowcolor{gray!30}

\multicolumn{24}{c}{\textbf{Proprietary MLLMs}} \\
\rowcolor{gray!15}
Gemini 1.5 pro~\citep{team2024gemini} & 1 fps
& 79.02 & 80.47 & 83.54 & 79.67 & 80.00 & 84.74 & 77.78 & 64.23 & 71.95 & 48.70 & 75.69
& 46.80 & 39.60 & 74.90 & 80.00 & 60.22
& 51.41 & 40.73 & 54.80 & 45.10 & 48.73
& 67.07 \\
\rowcolor{gray!15}
GPT-4o~\citep{hurst2024gpt} & 64
& 77.11 & 80.47 & 83.91 & 76.47 & 70.19 & 83.80 & 66.67 & 62.19 & 69.12 & 49.22 & 73.28
& 41.20 & 37.20 & 43.60 & 56.00 & 44.50
& 41.20 & 38.40 & 32.80 & 56.86 & 38.70
& 60.15 \\
\rowcolor{gray!15}
Claude 3.5 Sonnet~\citep{anthropic2024claude35sonnet}  & 20
& 80.49 & 77.34 & 82.02 & 81.73 & 72.33 & 75.39 & 61.11 & 61.79 & 69.32 & 43.09 & 72.44
& 31.60 & 34.00 & 32.80 & 48.80 & 36.80
& 38.40 & 34.80 & 34.40 & 64.71 & 37.70
& 57.68 \\
\midrule

\multicolumn{24}{c}{\textbf{Open-Source Offline Video MLLMs}} \\
LLaVA-OV-7B~\citep{li2024llava} & 32
& 80.38 & 74.22 & 76.03 & 80.72 & 72.67 & 71.65 & 67.59 & 65.45 & 65.72 & 45.08 & 71.12
& 40.80 & \textbf{37.20} & 33.60 & 44.80 & 38.40
& 35.60 & 36.00 & 27.27 & 29.55 & 32.74
& 56.36 \\
Qwen2-VL-7B~\citep{yang2024qwen2} & 0.2--1 fps$^{\dagger}$
& 75.20 & \underline{82.81} & 73.19 & 77.45 & 68.32 & 71.03 & 72.22 & 61.19 & 61.47 & 46.11 & 69.04
& 41.20 & 22.00 & 32.80 & 43.60 & 34.90
& 31.20 & 26.00 & 39.60 & 22.73 & 31.66
& 54.14 \\
MiniCPM-V 2.6-8B~\citep{yao2024minicpm}  & 32
& 71.93 & 71.09 & 77.92 & 75.82 & 64.60 & 65.73 & 70.37 & 56.10 & 62.32 & 53.37 & 67.44
& 40.80 & 24.00 & 34.00 & 41.20 & 35.00
& 34.00 & 31.60 & 41.92 & 22.22 & 34.97
& 53.85 \\
InternVL-V2-8B~\citep{chen2024far}  & 16
& 68.12 & 60.94 & 69.40 & 77.12 & 67.70 & 62.93 & 59.26 & 53.25 & 54.96 & 56.48 & 63.72
& 37.60 & 26.40 & 37.20 & 42.00 & 35.80
& 32.00 & 31.20 & 32.32 & 40.91 & 32.42
& 51.40 \\
Kangaroo-7B~\citep{liu2026kangaroo}  & 64
& 71.12 & \textbf{84.38} & 70.66 & 73.20 & 67.08 & 61.68 & 56.48 & 55.69 & 62.04 & 38.86 & 64.60
& 37.60 & 31.20 & 28.80 & 39.20 & 34.20
& 32.80 & 26.40 & 33.84 & 16.00 & 30.06
& 51.10 \\
LongVA-7B~\citep{zhang2024long} & 128
& 70.03 & 63.28 & 61.20 & 70.92 & 62.73 & 59.50 & 61.11 & 53.66 & 54.67 & 34.72 & 59.96
& 39.60 & 32.40 & 28.00 & 41.60 & 35.40
& 32.80 & 29.60 & 30.30 & 15.91 & 29.95
& 48.66 \\
VILA-1.5-8B~\citep{lin2024vila}  & 14
& 53.68 & 49.22 & 70.98 & 56.86 & 53.42 & 53.89 & 54.63 & 48.78 & 50.14 & 17.62 & 52.32
& 41.60 & 26.40 & 28.40 & 36.00 & 33.10
& 26.80 & 34.00 & 23.23 & 17.65 & 27.35
& 43.20 \\
Video-CCAM-14B~\citep{fei2024video} & 96
& 56.40 & 57.81 & 65.30 & 62.75 & 64.60 & 51.40 & 42.59 & 47.97 & 49.58 & 31.61 & 53.96
& 33.60 & 22.00 & 28.40 & 34.80 & 29.70
& 27.60 & 24.40 & 16.67 & 22.73 & 22.88
& 42.53 \\
Video-LLaMA2-7B~\citep{cheng2024videollama} & 32
& 55.86 & 55.47 & 57.41 & 58.17 & 52.80 & 43.61 & 39.81 & 42.68 & 45.61 & 35.23 & 49.52
& 30.40 & 32.40 & 30.40 & 36.00 & 32.40
& 24.80 & 26.80 & 18.67 & 0.00 & 21.93
& 40.40 \\
\midrule

\multicolumn{24}{c}{\textbf{Open-Source Online Video MLLMs}} \\
Flash-VStream-7B~\citep{zhang2025flash}  & -
& 25.89 & 43.57 & 24.91 & 23.87 & 27.33 & 13.08 & 18.52 & 25.20 & 23.87 & 48.70 & 23.23
& 25.91 & 24.90 & 25.60 & 28.40 & 26.00
& 24.80 & 25.20 & 26.80 & 1.96 & 24.12
& 24.04 \\
VideoLLM-online-8B~\citep{chen2024videollm}  & 2 fps
& 39.07 & 40.06 & 34.49 & 31.05 & 45.96 & 32.40 & 31.48 & 34.16 & 42.49 & 27.89 & 35.99
& 31.20 & 26.51 & 24.10 & 32.00 & 28.45
& 24.19 & 29.20 & 30.80 & 3.92 & 26.55
& 32.48 \\
Dispider-7B~\citep{qian2025dispider} & 1 fps
& 74.92 & 75.53 & 74.10 & 73.08 & 74.44 & 59.92 & 76.14 & 62.91 & 62.16 & 45.80 & 67.63
& 35.46 & 25.26 & 38.57 & 43.34 & 35.66
& 39.62 & 27.65 & 34.80 & 25.34 & 33.61
& 53.12 \\
TimeChat-Online-7B~\citep{yao2025timechat}& 1 fps
& 80.22 & 82.03 & 79.50 & 83.33 & 76.10 & 78.50 & 78.70 & 64.63 & 69.60 & \underline{57.98} & 75.36
& -- & -- & -- & -- & --
& -- & -- & -- & -- & --
& -- \\
QueryStream-7B~\citep{zhang2026querystream} & 1 fps
& \textbf{82.38} & \textbf{84.38} & 79.18 & 82.37 & \underline{77.99} & 81.31 & 78.70 & 65.04 & 69.32 & 47.34 & 75.32
& -- & -- & -- & -- & --
& -- & -- & -- & -- & --
& -- \\
\midrule
InternVL3.5-8B~\citep{wang2025internvl3}  & 1fps
& 78.86 & \underline{82.81} & 85.17 & 81.41 & \underline{77.99} & 79.75 & 83.33 & \underline{70.73} & \underline{76.99} & 55.32 & 77.60
& 40.40 & 30.80 & 42.80 & \underline{53.60} & 41.90
& 42.40 & 36.40 & \textbf{48.40} & 40.00 & 41.80
& 61.71 \\
\rowcolor{softblue}
\quad + \methodname  & 1fps
& \underline{82.11} & \underline{82.81} & \textbf{89.27} & \underline{84.62} & \textbf{83.65} & \underline{87.54} & \underline{87.04} & \textbf{73.17} & \textbf{77.56} & 53.72 & \textbf{80.72}
& \textbf{48.40} & \underline{34.40} & \textbf{48.00} & 51.20 & \textbf{45.50}
& \textbf{47.60} & \textbf{44.80} & 46.80 & \underline{53.23} & \underline{48.10}
& \textbf{65.64} \\
Qwen3-VL-8B~\citep{Qwen3-VL}  & 1fps
& 75.10 & 76.60 & 80.10 & 81.10 & 77.40 & 71.70 & 77.80 & 61.40 & 69.30 & 55.90 & 72.80
& 41.20 & 32.00 & \underline{46.80} & \textbf{55.20} & \underline{43.80}
& 40.00 & 31.60 & 34.40 & 52.40 & 36.75
& 59.33 \\
\rowcolor{softblue}
\quad + \methodname  & 1fps
& 81.84 & 79.69 & \underline{87.70} & \textbf{85.58} & 76.73 & \textbf{90.97} & \textbf{87.96} & 69.11 & 75.28 & \textbf{59.04} & \underline{80.16}
& \underline{43.20} & 26.40 & 44.00 & 49.20 & 40.70
& \underline{44.40} & \underline{41.60} & \underline{48.00} & \textbf{62.90} & \textbf{49.20}
& \underline{64.51} \\
\bottomrule
\end{tabular}%
}
\vspace{-2mm}
\label{tab:bench_sb}
\end{table*}

\subsection{Comparison with State-of-the-art Methods}
\label{subsec:sota}
Tables~\ref{tab:bench_ovo} and~\ref{tab:bench_sb} present results on OVO-Bench and StreamingBench, respectively. \methodname{} achieves the best overall performance among open-source models on both benchmarks, consistently improving both Qwen3-VL (51.70$\to$58.02 on OVO-Bench; 59.33$\to$64.51 on StreamingBench) and InternVL3.5 (47.84$\to$49.03; 61.71$\to$65.64). The gains align with our two design goals: the Time Dictionary improves detail-sensitive tasks (e.g., OCR, TR) where fine-grained cues must be preserved, while temporal strategy planning benefits temporally grounded tasks (e.g., EPM, PO) that require selecting evidence from the right temporal scope. A per-category analysis is provided in Appendix~\ref{app:per_category}. With the same Qwen2.5-VL-7B backbone, \methodname{} also exceeds QueryStream and StreamAgent by 5.01 and 3.12 points overall on OVO-Bench (Appendix~\ref{app:same_backbone}).

\begin{table}[t!]
\centering
\caption{\textbf{Comparison of memory representations under an equal token budget on OVO-Bench.} \#Tokens denotes the average number of memory tokens injected into the answer model at inference time. Qwen3-VL-8B is used as the backbone.}
\vspace{2mm}
\resizebox{0.95\linewidth}{!}{
\begin{tabular}{@{}l c c cc c c@{}}
\toprule
\textbf{Memory Type} & \textbf{\#Tokens}  & \textbf{Real-Time Acc.} & \textbf{Backward Acc.} & \textbf{Forward Acc.} & \textbf{Overall Acc.}  \\
\midrule
No Memory & -- & 63.24 & 43.44&48.42&51.70  \\
Visual Prototype Memory&203.11 & 66.19 & 44.16 & \textbf{ 51.36}& 53.90\\
Flat Text Memory& 236.60 & 70.98 & 48.57 & 47.83&  55.79 \\
\textbf{Structured Dictionary (Ours)}&245.23 &\textbf{73.28} & \textbf{51.37} & 49.41 & \textbf{58.02} \\
\bottomrule
\end{tabular}
}
\vspace{-4mm}
\label{tab:text_vs_visual}
\end{table}

\vspace{-1mm}
\subsection{Effect of \methodname{}}
\label{effect_td}

\noindent\textbf{Effect of Structured Time Dictionary.}
\label{subsec:effect_of_memory_type}
Table~\ref{tab:text_vs_visual} compares alternative memory representations under a similar memory-token budget on queries that require recovering fine-grained past visual details. We compare three memory designs: \textit{visual prototype memory}, which clusters accumulated visual tokens into $K$ representative prototypes via $k$-means over the streaming sequence ($K{=}2$, configured to match the token budget of our method for fair comparison); \textit{flat text memory}, which stores a single unstructured free-form caption per frame; and our \textit{structured dictionary}, which decomposes each caption into eight semantic fields. Relative to the no-memory baseline, both visual prototype memory and flat text memory improve performance, confirming that retaining historical information is essential for streaming video QA. However, the proposed structured dictionary achieves the strongest overall result, improving the direct Qwen3-VL baseline from 51.70 to 58.02 and outperforming flat text memory by 2.23 points overall, with consistent gains in both real-time and backward accuracy. This suggests that the benefit does not come merely from storing more past content, but from organizing memory into semantically typed fields that preserve fine-grained details while making relevant evidence easier to retrieve. 

\begin{table*}[t]
\centering
\begin{minipage}[t]{0.42\textwidth}
\vspace{0pt}
\centering
\caption{\textbf{Effect of temporal routing on OVO-Bench Backward Tracing.} EPM: Episodic Memory, ASI: Action Sequence Identification, HLD: Hallucination Detection. $^{\dagger}$Average over five runs.}
\vspace{2mm}
\label{tab:routing_effect}
\resizebox{\linewidth}{!}{
\begin{tabular}{@{}l c c c c@{}}
\toprule
\textbf{Setting} & \multicolumn{4}{c}{\textbf{Backward Tracing}} \\
\cmidrule(lr){2-5}
& EPM & HLD & ASI & Avg. \\
\midrule
Baseline (no memory)~\citep{Qwen3-VL} & 52.53 & 10.22 & 67.57 & 43.44 \\
\midrule
All-to-Past Lookback & 57.91 & \textbf{25.81} & 67.57 & 50.43 \\
Random Routing$^{\dagger}$ & 52.12 & 19.78 & 63.24 & 45.05 \\
Our Routing & \textbf{59.26} & 25.27 & \textbf{69.59} & \textbf{51.37} \\
\bottomrule
\end{tabular}
}
\end{minipage}%
\hfill
\begin{minipage}[t]{0.56\textwidth}
\vspace{0pt}
\centering
\caption{\textbf{Schema-group ablation for the Time Dictionary.} Each row removes one of four schema groups while keeping all else fixed. Overall is computed only over the listed subtasks, not over the full benchmark.}
\vspace{2mm}
\label{tab:ablation_schema_groups}
\resizebox{\linewidth}{!}{
\begin{tabular}{l ccc ccc}
\toprule
\multirow{2}{*}{Setting} & \multicolumn{3}{c}{OVO-Bench} & \multicolumn{3}{c}{StreamingBench} \\
\cmidrule(lr){2-4} \cmidrule(lr){5-7}
 & OCR & OJR & Overall & TR & SU & Overall \\
\midrule
Baseline~\citep{Qwen3-VL} & 75.17 & 57.07 & 66.12 & 71.70 & 61.40 & 67.23 \\
Full Schema (Ours) & 87.25 & 74.46 & 80.86 & 90.97 & 69.11 & 81.48 \\
\midrule
w/o Background & 83.89 & 69.57 & 76.73 & 90.97 & 69.11 & 81.48 \\
w/o Entity-Centric & 83.89 & 72.83 & 78.36 & 91.59 & 67.07 & 80.95 \\
w/o Explicit Text & 83.22 & 73.37 & 78.30 & 85.98 & 68.29 & 78.31 \\
w/o Contextual Sent. & 85.23 & 71.74 & 78.49 & 89.41 & 67.07 & 79.72 \\
\bottomrule
\end{tabular}
}
\end{minipage}
\end{table*}

\noindent\textbf{Effect of Temporal Strategy Planning.}
\label{subsec:effect_of_selection}
Table~\ref{tab:routing_effect} provides direct evidence for temporal strategy planning. While all memory-based settings outperform the no-memory baseline, our routing achieves the best backward-tracing average (51.37) with top scores on EPM (59.26) and ASI (69.59). Notably, indiscriminate lookback is not universally beneficial: some backward-tracing questions still depend on current grounding, and always retrieving past evidence can introduce temporally irrelevant distractors. Query-conditioned routing preserves the benefit of memory retrieval while avoiding unnecessary access to irrelevant history, confirming that streaming VQA requires selecting the appropriate temporal strategy per query, not just retaining past evidence. Here, Random Routing assigns each query to the past, current, or future strategy uniformly at random instead of using the planner; other routing variants are reported in Appendix~\ref{app:random_routing}.

\subsection{Diagnostic Analysis of \methodname{}}
\label{subsec:reliability}

\begin{table}[t]
\centering
\begin{minipage}[t]{0.48\linewidth}
\centering
\caption{\textbf{Confusion matrix of the Temporal Strategy Planner on StreamingBench.} The 150 queries are sampled 50 each from Real-Time, Contextual, and Proactive Output tasks. Rows: human-assigned temporal label; columns: planner prediction.}
\label{tab:planner_cm}
\vspace{2mm}
\resizebox{\linewidth}{!}{
\begin{tabular}{@{}l ccc c@{}}
\toprule
\textbf{Human} & Current & Past & Future & Recall \\
\midrule
Current ($n{=}70$) & \textbf{47} & 23 & 0 & 67.1 \\
Past ($n{=}30$) & 6 & \textbf{24} & 0 & 80.0 \\
Future ($n{=}50$) & 0 & 2 & \textbf{48} & 96.0 \\
\midrule
\multicolumn{5}{@{}l}{Overall agreement: 79.3\% (GPT-4o 86.7\%, GPT-5 85.3\%)} \\
\bottomrule
\end{tabular}
}
\end{minipage}%
\hfill
\begin{minipage}[t]{0.48\linewidth}
\centering
\caption{\textbf{Effect of the number of visual frames given to the answer model on OVO-Bench.} Qwen3-VL-8B is used as the answer model, and Time Dict.\ \cmark{} denotes the full \methodname{} framework.}
\label{tab:visual_input}
\vspace{2mm}
\resizebox{\linewidth}{!}{
\begin{tabular}{@{}l c cccc@{}}
\toprule
\textbf{Frames} & \textbf{Time Dict.} & \textbf{RT} & \textbf{BW} & \textbf{FWD} & \textbf{Overall} \\
\midrule
512 (baseline) & \xmark & 63.24 & 43.44 & 48.42 & 51.70 \\
512 (Ours) & \cmark & 73.28 & 51.37 & \textbf{49.41} & \textbf{58.02} \\
1 & \cmark & \textbf{77.78} & 52.14 & 41.50 & 57.14 \\
0 & \cmark & 68.94 & \textbf{55.47} & 34.84 & 53.08 \\
\bottomrule
\end{tabular}
}
\end{minipage}
\vspace{-3mm}
\end{table}

\noindent\textbf{Planner Accuracy.} On 150 human-labeled StreamingBench queries (Appendix~\ref{app:annotation}), the planner agrees with the human labels on 79.3\% of queries (GPT-4o: 86.7\%, GPT-5: 85.3\%), and its future predictions reach 100\% precision and 96\% recall (Table~\ref{tab:planner_cm}). Most disagreements (23 of 31) route a current query to past, which may add irrelevant history, while errors that omit the required evidence are rare (8 of 150). Replacing the planner's routes with the human labels improves accuracy on this subset from 58.7\% to 61.3\%, indicating a limited downstream impact of routing errors.

\noindent\textbf{Evidence Coverage.} We randomly sampled 100 past-routed OVO Backward queries and manually inspected whether the answer-critical evidence appears in the selected captions, in their source frames, or elsewhere in the Time Dictionary (details in Appendix~\ref{app:annotation}). The selected captions contain the evidence for 48\% of the queries, and the dictionary contains it for 65\% in total. For 10\%, the evidence is visible in a selected source frame but missing from its caption, and for 17\%, it is stored in the dictionary but not retrieved. For the remaining 25\%, no supporting caption is found; since only the source frames of the selected captions were inspected, these cases may stem from the 1-fps sampling or from omissions in captions of non-selected frames.

\noindent\textbf{Multi-moment Retrieval.} Past lookback provides the selected captions jointly in temporal order and often covers multiple moments. For Forward instances routed to past, it retrieves 2.82 captions on average and up to 19; among the 619 instances with $K^{\star}\ge2$, the retrieved captions span a median of 6\,s and a 90th percentile of 19\,s, with 28.3\% spanning more than 10\,s.

\noindent\textbf{Effect of Visual Frames.} Reducing the visual input given to the answer model from 512 frames to a single frame lowers the overall score from 58.02 to 57.14 (Table~\ref{tab:visual_input}), mostly on Forward questions that involve ongoing events (49.41$\to$41.50). Notably, even without any frame, \methodname{} still outperforms the 512-frame baseline (53.08 vs.\ 51.70) and attains the highest Backward accuracy (55.47), showing that the Time Dictionary by itself serves as an effective memory of the stream.

\subsection{Ablation Study on Time Dictionary Schema}
\label{subsec:ablation}

Table~\ref{tab:ablation_schema_groups} reports a schema-group ablation that removes one semantic field group at a time while keeping all other components unchanged. The full schema achieves the best overall performance on both benchmarks. Removing \textit{Explicit Text} causes the largest drop on Text-Rich Understanding (TR: 90.97 $\rightarrow$ 85.98) and Optical Character Recognition (OCR: 87.25 $\rightarrow$ 83.22), confirming that fine-grained on-screen cues such as signs, logos, and labels require a dedicated field to remain recoverable at query time. Removing \textit{Entity-Centric Dynamics} (Objects + People + Actions + Spatial) degrades OCR (87.25 $\rightarrow$ 83.89), Object Recognition (OJR: 74.46 $\rightarrow$ 69.57), and Spatial Understanding (SU: 69.11 $\rightarrow$ 67.07), showing that object-level attributes and inter-entity relations likewise benefit from explicit decomposition. The remaining two groups (Background, Contextual Sentence) also contribute to overall performance, but their removal patterns are less task-specific; we attribute this to the fact that the eight schema fields are not fully independent, and the model leverages overlapping cues across fields to partially compensate for a missing group. Altogether, these results support our design: structured decomposition into typed fields preserves fine-grained evidence that would otherwise be lost in a flat or compressed representation.

\subsection{Asynchronous Streaming and Query Latency}
\label{subsec:discussion}

Streaming-state updates and query answering are executed asynchronously and therefore should not be collapsed into a single end-to-end latency number. Under a 1 fps input stream, maintaining the Time Dictionary requires 637.51 ms per second of video, which stays within the real-time budget and leaves headroom for continuous online operation. Once this state has been prepared, query-time reasoning introduces only modest overhead over direct visual QA: the total latency is 574.44 ms for past queries, 557.62 ms for current queries, and 565.11 ms for future queries with one polling step, compared with 547.22 ms for the visual-only baseline, corresponding to only 10--27 ms of additional delay depending on the temporal branch. This breakdown reflects the practical advantage of our design: the heavier video processing cost is amortized during streaming, while user-time inference operates on compact structured state, allowing responsive query answering without reprocessing the full visual history for each question. A detailed latency breakdown is provided in Table~\ref{tab:latency_async}. Scalability beyond 1 fps and a comparison with InfiniPot-V~\citep{kim2025infinipot} are provided in Appendices~\ref{app:scalability} and~\ref{app:infinipot}.

\section{Discussion and Conclusion}\label{sec:conclusion}

\paragraph{Limitation.} TimeTraveler relies on caption-based textual evidence as the persistent memory representation. While this improves query recoverability and efficiency, it may inherit errors from the captioner, especially for small objects, ambiguous actions, or OCR under low resolution; answer-critical details omitted from a caption cannot be recovered from the dictionary (10\% of the sampled queries in Section~\ref{subsec:reliability}). The current planner and future gate are prompt-based frozen models, so misclassification of temporal scope can lead to suboptimal evidence selection. Further limitations and future directions are discussed in Appendix~\ref{app:limitations}.

\paragraph{Conclusion.} We present \methodname{}, a streaming video understanding framework built on two complementary ideas: storing past observations in an explicit and queryable structure, and matching each query to the appropriate temporal scope. \methodname{} maintains a lightweight Time Dictionary that records each observed moment as structured textual evidence, preserving fine-grained cues in a more recoverable form. On top of this memory, a Temporal Strategy Planner determines whether the answer should be grounded in past observations, the current scene, or future evidence, enabling query-specific evidence access. Extensive experiments and analyses on OVO-Bench and StreamingBench show that this combination improves both detail-sensitive and temporally grounded reasoning, while maintaining practical efficiency under asynchronous streaming deployment.

\section*{Acknowledgments}
This work was supported in part by the National Research Foundation of Korea (NRF) grant funded by the Korea government (MSIT) (No. RS-2024-00334321), by the MSIT (Ministry of Science and ICT), Korea, under the ITRC (Information Technology Research Center) support program (No. IITP-2026-RS-2023-00258649) supervised by the IITP (Institute for Information and Communications Technology Planning and Evaluation), and by Global - Learning \& Academic research institution for Master’s·PhD students, and Postdocs(G-LAMP) Program of the National Research Foundation of Korea(NRF) grant funded by the Ministry of Education(No. RS-2025-25442355), by Institute of Information and Communications Technology Planning and Evaluation (IITP) grant funded by the Korea government (MSIT) (No. RS-2022-II220078, RS-2025-02653113).

{\small
\bibliographystyle{unsrt}
\bibliography{main}
}

\clearpage

\section*{Appendix Contents}

\startcontents[appendices]
\printcontents[appendices]{}{1}{}

\appendix

\newpage

\section{Additional Analysis of \methodname }

\begin{table*}[t!]
\centering
\caption{\textbf{Asynchronous latency breakdown of \methodname{}.} The two blocks run independently: stream processing executes continuously, while query processing is triggered on demand. Stream latency is normalized to one second of input at 1\,fps. For the past path, the QA cost is linearly interpolated between the measured 1- and 8-caption settings at the average number of retrieved captions ($\bar{K}^{\star}{=}2.68$ over past-routed OVO-Bench queries). Qwen3-VL-8B is used as the backbone.}
\label{tab:latency_async}
\vspace{0.3em}
\begin{minipage}[t]{0.32\textwidth}
\centering
\scalebox{0.82}{
\begin{tabular}{@{}l c c@{}}
\toprule
\rowcolor{gray!30}
\multicolumn{3}{c}{\textbf{Stream Processing} {\small (always-on)}} \\
\midrule
\textbf{Stage} & \textbf{Latency} & \textbf{TFLOPs} \\
\midrule
Video decode & 15.72 ms & -- \\
Captioner & 605.82 ms & 1.53 \\
Embedder & 15.97 ms & 0.20 \\
\midrule
\textbf{Total (per 1\,s)} & \textbf{637.51 ms} & \textbf{1.73} \\
\bottomrule
\end{tabular}
}
\end{minipage}%
\hfill
\begin{minipage}[t]{0.66\textwidth}
\centering
\scalebox{0.82}{
\begin{tabular}{@{}l l c c@{}}
\toprule
\rowcolor{gray!30}
\multicolumn{4}{c}{\textbf{Query Processing} {\small (on-demand)}} \\
\midrule
\textbf{Path} & \textbf{Stage} & \textbf{Latency} & \textbf{TFLOPs} \\
\midrule
Baseline & QA (Visual-only) & 547.22 ms & 1.42 \\
\midrule
\multirow{4}{*}{Our Past}
& Scope planning & 9.60 ms & 1.39 \\
& Retrieval & 15.62 ms & 0.03 \\
& QA (2.68 cap) & 549.22 ms & 8.28 \\
& \textbf{Total} & \textbf{574.44 ms} & \textbf{9.70} \\
\midrule
\multirow{3}{*}{Our Current}
& Scope planning & 9.60 ms & 1.39 \\
& QA (1 cap) & 548.02 ms & 3.81 \\
& \textbf{Total} & \textbf{557.62 ms} & \textbf{5.2} \\
\midrule
\multirow{4}{*}{Our Future}
& Scope planning & 9.60 ms & 1.39 \\
& Gating ($\times$1; +7.49\,ms/extra) & 7.49 ms & 5.73 \\
& QA (1 cap) & 548.02 ms & 3.81 \\
& \textbf{Total} & \textbf{565.11 ms} & \textbf{10.93} \\
\bottomrule
\end{tabular}
}
\end{minipage}
\end{table*}

\begin{table}[t]
\centering
\caption{\textbf{Memory footprint comparison across memory designs on Qwen3-VL-8B~\citep{Qwen3-VL}.} Visual tokens per frame are measured at 240 for inputs resized to $\texttt{max\_pixels}=401408$. Visual prototype uses $K{=}32$ slots. Our structured dictionary stores 800-byte UTF-8 caption text and a 1024-dim FP16 embedding per frame. Retrieval embeddings are maintained in a finite queue covering the most recent 600\,s, so their cost is capped beyond 10 minutes.}
\label{tab:memory_footprint}
\vspace{2mm}
\resizebox{\textwidth}{!}{%
\begin{tabular}{l c | c c c c c}
\toprule
Memory Design & Per-frame & 1\,min & 5\,min & 10\,min & 30\,min & 1\,hour \\
\midrule
Visual Full KV Cache & 33.75\,MB & 1.98\,GB & 9.89\,GB & 19.78\,GB & 59.33\,GB & 118.65\,GB \\
Visual Hidden State & 1.88\,MB & 112.5\,MB & 562.5\,MB & 1.10\,GB & 3.30\,GB & 6.59\,GB \\
Visual Prototype ($K{=}32$) & -- & 1.6\,MB & 1.6\,MB & 1.6\,MB & 1.6\,MB & 1.6\,MB \\
\midrule
\multicolumn{7}{l}{\textit{Time Dictionary (Ours)}} \\
\quad Caption Text & 0.78\,KB & 46\,KB & 234\,KB & 469\,KB & 1.4\,MB & 2.8\,MB \\
\quad Retrieval Embedding & 2.00\,KB & 120\,KB & 600\,KB & 1.2\,MB & 1.2\,MB & 1.2\,MB \\
\rowcolor{gray!10}
\quad Total & 2.78\,KB & 167\,KB & 834\,KB & 1.6\,MB & 2.6\,MB & 4.0\,MB \\
\bottomrule
\end{tabular}%
}
\end{table}

\subsection{Memory footprint}
\label{memory size}
Table~\ref{tab:memory_footprint} quantifies the storage cost that underlies our design choice of textual over visual memory. A central motivation of \methodname{} is that streaming memory should remain both \emph{fine-grained} and \emph{lightweight}; visual representations can preserve perceptual detail but scale linearly with stream length at a prohibitive rate, while compressed prototypes($K{=}32$ slots, constructed via $k$-means clustering of visual tokens) stay compact but sacrifice queryable detail. Our structured dictionary occupies only 2.78\,KB per frame---roughly $690\times$ less than a visual hidden state and over $12{,}000\times$ less than a full KV cache---yet retains semantically typed, directly queryable evidence. Over a one-hour stream at 1\,FPS, the entire dictionary amounts to 4.0\,MB, whereas visual KV memory would require over 118\,GB. Even the most compact visual baseline, prototype memory with $K{=}32$ slots, achieves its small footprint only by collapsing the stream into a fixed-size state that discards temporal indexing and fine-grained cues. In contrast, our dictionary scales linearly but remains practically negligible, keeping every observed moment individually addressable while adding virtually no system-level memory burden.

\subsection{Scalability beyond 1 fps.}
\label{app:scalability}
\begin{table}[t]
\centering
\caption{\textbf{Accuracy--throughput trade-off of the captioning stage.} Captioning throughput is measured in frames per second; all other components are identical to our default setting with Qwen3-VL-8B as the answer model. OVO-Bench reports the three category averages and the overall score; SB denotes StreamingBench overall.}
\label{tab:scalability}
\vspace{2mm}
\resizebox{\linewidth}{!}{
\begin{tabular}{@{}l c c cccc c@{}}
\toprule
\textbf{Captioner} & \textbf{Frames/call} & \textbf{Throughput} & \textbf{RT} & \textbf{BW} & \textbf{FWD} & \textbf{OVO} & \textbf{SB} \\
\midrule
No memory (baseline) & -- & -- & 63.24 & 43.44 & 48.42 & 51.70 & 59.33 \\
\midrule
Qwen3-VL-2B-FP8 (Ours) & 1 & 1.65 & \textbf{73.28} & \textbf{51.37} & 49.41 & \textbf{58.02} & \textbf{64.51} \\
Qwen3.5-0.8B & 1 & 3.52 & 72.46 & 50.21 & \textbf{50.94} & 57.87 & 62.92 \\
Qwen3-VL-2B-FP8 & 8 & \textbf{8.97} & 69.31 & 48.97 & 50.91 & 56.40 & 61.55 \\
\bottomrule
\end{tabular}
}
\end{table}

We follow the 1-fps setting of prior streaming methods~\citep{zhang2025flash,zhang2026querystream,qian2025dispider,yao2025timechat}. Captioning dominates stream processing (605.82\,ms per frame), and the dictionary update leaves about 362\,ms of headroom per second at 1 fps; since state updates run asynchronously, they do not block answer generation. For higher input rates, two complementary axes extend the operating range on the same eight-field schema (Table~\ref{tab:scalability}): clip-level dictionary captioning, which writes one structured caption per clip of consecutive frames and thus reduces the captioning frequency by the clip length, and a lighter captioner (Qwen3.5-0.8B) under the identical prompt and schema. The 0.8B captioner raises captioning throughput from 1.65 to 3.52 frames/s while staying within 0.15 points of our default overall OVO-Bench score, and 8-frame clips sustain 8.97 frames/s while retaining gains of 4.70 (OVO-Bench) and 2.22 (StreamingBench) points over the backbone. The two axes can be combined: increasing the clip size from 4 to 32 frames raises the sustainable input rate from about 5 to 33 fps, and the 0.8B captioner with 8-frame clips reaches about 22 fps.

\subsection{Comparison with memory-efficient streaming.}
\label{app:infinipot}
\begin{table*}[t]
\centering
\caption{\textbf{Accuracy--efficiency comparison with InfiniPot-V~\citep{kim2025infinipot}.} Both methods use the same backbone (Qwen2.5-VL-7B), environment, and evaluation protocol, and InfiniPot-V is run with its official implementation. Memory budget counts the answer-time tokens provided to the answer model. $^{\dagger}$InfiniPot-V reuses its compressed KV states built during streaming, so its answer latency mainly consists of question processing and answer generation, whereas \methodname{} prefills its visual and textual evidence at query time.}
\label{tab:infinipot}
\vspace{2mm}
\resizebox{\textwidth}{!}{\footnotesize
\setlength{\tabcolsep}{4pt}
\newcommand{\tc}[2]{\begin{tabular}[c]{@{}c@{}}#1\\#2\end{tabular}}
\begin{tabular}{@{}l l l c c c c c c@{}}
\toprule
\textbf{Method} & \textbf{Visual memory} & \tc{\textbf{Text}}{\textbf{memory}} & \tc{\textbf{Memory}}{\textbf{budget}} & \tc{\textbf{Memory}}{\textbf{footprint}} & \tc{\textbf{Answer}}{\textbf{latency}} & \tc{\textbf{Answer-model}}{\textbf{memory}} & \textbf{OVO-BW} & \textbf{SB-RT} \\
\midrule
InfiniPot-V & 4K compressed KV & -- & 3.83K & 220\,MB & \textbf{53.75\,ms}$^{\dagger}$ & 16.80\,GB & 45.60 & 76.76 \\
\midrule
\multirow{3}{*}{\methodname{}} & Current frame & $K^{\star}\!\approx\!2.5$ & \textbf{0.70K} & \textbf{39\,MB} & 58.79\,ms & \textbf{16.62\,GB} & \textbf{57.14} & 69.96 \\
 & Current frame & Top-20 & $\sim$3.8K & 214\,MB & 72.02\,ms & 16.79\,GB & 56.37 & 70.88 \\
 & Uniformly selected frames & $K^{\star}\!\approx\!2.5$ & $\sim$3.8K & 214\,MB & 91.81\,ms & 16.79\,GB & 53.39 & \textbf{77.72} \\
\bottomrule
\end{tabular}
}
\end{table*}

We compare with InfiniPot-V~\citep{kim2025infinipot}, which bounds the visual KV cache, using the same backbone (Qwen2.5-VL-7B), environment, and protocol (Table~\ref{tab:infinipot}). Under a comparable answer-time budget of about 4K tokens, \methodname{} improves OVO Backward accuracy over InfiniPot-V by 7.79 points with uniformly selected frames and by 10.77 points when the budget is spent on retrieved captions, and the one-frame setting reaches 57.14 with only 0.70K tokens and a 39\,MB footprint, suggesting that structured textual evidence is an effective representation for recovering historical evidence under a constrained budget. On StreamingBench Real-Time, the matched-budget setting with uniformly selected frames is competitive (77.72 vs.\ 76.76), whereas the text-heavy settings are lower (69.96--70.88), reflecting that real-time perception benefits more from visual frames than from historical text. InfiniPot-V attains lower answer latency because it reuses compressed KV states at query time, and our captioner/planner and embedder require about 4.66\,GB of additional resident GPU memory during streaming. Since InfiniPot-V bounds the visual KV cache while \methodname{} adds lightweight structured text entries and retrieves a small query-relevant subset, the two approaches manage different parts of streaming memory and are potentially complementary.

\subsection{Retrieval horizon and index scaling.}
\label{app:horizon}
The 600-s horizon applies only to the retrieval keys (embeddings); caption values are lightweight (0.78\,KB per caption; Table~\ref{tab:memory_footprint}) and are kept permanently, so no evidence is discarded. Because keys accumulate at every frame and the end of a stream is unknown, we bound the key index by a fixed horizon. Extending the indexed horizon from 600\,s to one hour changes the total per-query retrieval latency from 13.79 to 13.80\,ms, while the total storage for a one-hour stream grows from 4.0\,MB (Table~\ref{tab:memory_footprint}) to about 10.1\,MB; query encoding alone takes 13.76\,ms, about 99.7\% of the one-hour retrieval latency. Even with a four-hour index of 14,400 frame entries, cosine search takes only 0.083\,ms and the complete store occupies about 40.49\,MB. In a diagnostic run with a longer indexed horizon, evidence older than 600\,s was selected in only 3 of the 1,473 past-routed queries (0.2\%), so the cap was rarely binding in our setting. All results in this paper use the 600-s horizon; the 30-s lookahead $T_{\max}$ can likewise be extended when longer future observation is required.

\subsection{Same-backbone comparison.}
\label{app:same_backbone}
\begin{table}[t]
\centering
\caption{\textbf{Same-backbone comparison on OVO-Bench with Qwen2.5-VL-7B as the answer model.} $^{*}$Results reported in the original papers, which do not report backbone-only OVO-Bench scores; the backbone-only and \methodname{} rows are evaluated in our environment with the captioner, planner, embedder, and all hyperparameters unchanged.}
\label{tab:same_backbone}
\vspace{2mm}
{\small
\begin{tabular}{@{}l cccc@{}}
\toprule
\textbf{Method} & \textbf{Real-Time} & \textbf{Backward} & \textbf{Forward} & \textbf{Overall} \\
\midrule
QueryStream-7B$^{*}$~\citep{zhang2026querystream} & 61.40 & 42.10 & 39.03 & 47.51 \\
StreamAgent-7B$^{*}$~\citep{yang2025streamagent} & 61.30 & 41.70 & \textbf{45.40} & 49.40 \\
\midrule
Qwen2.5-VL-7B~\citep{bai2025qwen2} & 63.46 & 44.21 & 39.40 & 49.02 \\
\quad + \methodname{} & \textbf{71.71} & \textbf{48.96} & 36.90 & \textbf{52.52} \\
\bottomrule
\end{tabular}
}
\end{table}

The compared streaming methods use different answer backbones: Dispider-7B builds on Qwen2-7B, Flash-VStream-7B on Vicuna-7B, and QueryStream-7B and StreamAgent-7B on Qwen2.5-VL-7B. Table~\ref{tab:same_backbone} therefore evaluates \methodname{} with Qwen2.5-VL-7B, the answer backbone of QueryStream and StreamAgent, while keeping the captioner, planner, embedder, and hyperparameters unchanged. \methodname{} improves the backbone from 49.02 to 52.52 overall and exceeds QueryStream and StreamAgent by 5.01 and 3.12 points overall and by 10.31 and 10.41 points on Real-Time, indicating that the gains stem from the framework itself. Forward accuracy is lower with this backbone (39.40$\to$36.90), in line with the backbone-dependent Forward behavior in Appendix~\ref{app:per_category}.

\subsection{Per-category analysis.}
\label{app:per_category}
With Qwen3-VL, the largest gains appear in categories aligned with the intended role of each module: detail-sensitive categories supported by the structured \texttt{Text} and \texttt{Objects} fields, such as OCR (+12.08) and OJR (+17.39) on OVO-Bench and TR (+19.27) on StreamingBench (cf.\ Table~\ref{tab:ablation_schema_groups}), and temporally grounded categories supported by routing and gating, such as EPM (+6.73) and HLD (+15.05) on OVO-Bench and PO (+10.50) on StreamingBench (cf.\ Tables~\ref{tab:routing_effect} and~\ref{tab:future_strategy}). A few categories decrease. On OVO Forward, REC changes by $-$0.86 with Qwen3-VL and $-$10.45 with InternVL3.5, and SSR by $-$9.86 with InternVL3.5; since the deferred-observation gate is not used under the OVO multiple-triggering protocol (Appendix~\ref{app:ovo_protocol}), these changes likely reflect how each answer backbone uses retrieved evidence for counting and step recognition. On StreamingBench Omni-Source, the audio-dependent MA decreases (55.20$\to$49.20 with Qwen3-VL) because neither the Time Dictionary nor the answer model receives audio in our visual-only setting, whereas visually inferable categories such as ER improve with both backbones (41.20$\to$43.20 and 40.40$\to$48.40).

\subsection{Details of the planner and evidence-coverage evaluation.}
\label{app:annotation}
\noindent\textbf{Temporal routing.} We sampled 150 StreamingBench queries, 50 each from Real-Time Visual Understanding, Contextual Understanding, and Proactive Output; Proactive Output was included to cover future-directed queries, which OVO Forward does not provide under its multiple-triggering protocol. A human annotator assigned each query to current, past, or future using the question, options, ground-truth answer, query-time frame, and the corresponding caption, while remaining blind to the planner output. GPT-4o~\citep{hurst2024gpt} and GPT-5 were used as additional reference judges under the same routing criteria but without access to the ground-truth answer. To measure the downstream effect of routing errors, we replaced the planner predictions with the human-assigned routes on this subset.

\noindent\textbf{Ambiguous and mixed-temporal queries.} The planner outputs a single strategy, but the strategies are not mutually exclusive evidence sources: since the answer model always receives the query-time visual input $v$, the past route augments rather than replaces the current evidence, so queries that need both past and current evidence can be served by routing to past. Most remaining disagreements arise from such ambiguous past--current cases. Future intent is comparatively explicit in Proactive Output, which expresses it through trigger instructions; implicit future intent and queries that require combining past, current, and future evidence are not explicitly modeled by the single-route planner (Appendix~\ref{app:limitations}).

\noindent\textbf{Evidence coverage.} We randomly sampled 100 past-routed OVO Backward queries, including both correctly and incorrectly answered ones. For each query, the annotator inspected the question, options, ground-truth answer, the selected $K^{\star}$ captions, and their source frames, and assigned one of four outcomes: the evidence is contained in a selected caption (48\%); it is visible in a selected source frame but missing from its caption (10\%); it is absent from the selected captions but found elsewhere in the Time Dictionary (17\%); or it is not found in any dictionary caption (25\%).

\subsection{Random routing variants.}
\label{app:random_routing}
Random Routing replaces the Temporal Strategy Planner by assigning each query a temporal strategy uniformly at random and then applies the evidence acquisition of the selected strategy as usual.

\noindent\textbf{Random Routing.} Each query is assigned to the past, current, or future strategy, matching the action space of the planner; for the future strategy, the gate stops once sufficient evidence is available to answer the question. Over five runs, it achieves EPM 52.12$\pm$1.77, HLD 19.78$\pm$0.45, and ASI 63.24$\pm$1.13 on OVO-Bench Backward Tracing (mean values in Table~\ref{tab:routing_effect}), compared with 59.26, 25.27, and 69.59 for our routing; averaged over the three tasks, the scores are 45.05$\pm$0.63 and 51.37, respectively.

\noindent\textbf{Random Routing over current and past.} Each query is assigned to the current or past strategy only. When past is selected, the standard lookback is applied; when current is selected, the answer model receives the structured query-time caption. A single run achieves EPM 57.94, HLD 25.13, and ASI 66.52, and over five runs its task-averaged score is 50.01$\pm$0.70, compared with 51.37 for our routing under the same averaging.

Together, these results indicate that the gain of our routing stems from query-conditioned strategy selection rather than from mixing strategies.

\subsection{Effect of gating on proactive output task.}
Table~\ref{tab:future_strategy} examines the effect of future gating on StreamingBench Proactive Output. In this task, a question is posed before the answer-relevant event, and the model is evaluated at one second after the ground-truth moment on whether it correctly decides to answer and, if so, whether the answer is correct. The baseline makes this decision using only the current visual frames, while our settings additionally provide the Time Dictionary memory. Without gating, the 8B answer model itself judges whether to respond, but this actually hurts Qwen3-VL (52.40$\to$45.20), suggesting that the model struggles to make reliable respond-or-wait decisions when given the full memory context. With gating, a lightweight 2B model sees only the current-frame caption and the question to make the respond-or-wait decision, then hands off to the 8B model for answering. This yields substantial gains on both backbones (Qwen3-VL: +10.50; InternVL3.5: +13.23), confirming that separating the gating decision from the answering model is more effective.
\begin{table}[t]
\centering
\caption{\textbf{Effect of the future gating on StreamingBench Proactive Output.}}
\label{tab:future_strategy}
\vspace{2mm}
\scalebox{0.9}{
\begin{tabular}{@{}l l c@{}}
\toprule
\textbf{Backbone} & \textbf{Method} & \textbf{Acc.} \\
\midrule
\multirow{3}{*}{Qwen3-VL-8B}
& Baseline & 52.40 \\
& + Time Dictionary w/o gating & 45.20 \\
& + Time Dictionary w/ gating (Ours) & \textbf{62.90} \\
\midrule
\multirow{3}{*}{InternVL3.5-8B}
& Baseline & 40.00 \\
& + Time Dictionary w/o gating & 41.20 \\
& + Time Dictionary w/ gating (Ours) & \textbf{53.23} \\
\bottomrule
\end{tabular}
}
\end{table}

\subsection{Routing-based visual budget allocation.}
\label{app:visual_budget}
\begin{table}[t]
\centering
\caption{\textbf{Routing-based visual budget allocation on OVO-Bench.} Qwen3-VL-8B is used as the answer model, and the number of visual frames it receives is assigned per temporal strategy using the planner output.}
\label{tab:visual_budget}
\vspace{2mm}
{\small
\begin{tabular}{@{}l cccc@{}}
\toprule
\textbf{Policy} & \textbf{RT} & \textbf{BW} & \textbf{FWD} & \textbf{Overall} \\
\midrule
Ours (512 frames for all queries) & 73.28 & 51.37 & \textbf{49.41} & 58.02 \\
Past$\to$512, Current$\to$1 & \textbf{78.38} & 51.51 & 46.29 & \textbf{58.73} \\
Past$\to$0, Current$\to$1 & 76.34 & \textbf{54.99} & 38.92 & 56.75 \\
\bottomrule
\end{tabular}
}
\end{table}

Table~\ref{tab:visual_input} shows that Real-Time questions benefit most from a single current frame, Backward questions perform best when relying on the dictionary without frames, and Forward questions degrade as frames are removed. Motivated by this pattern, we assign different visual budgets to the past and current strategies using the planner output. Since the planner does not distinguish whether a past-routed query originates from a Backward or a Forward question, we evaluate two implementable policies (Table~\ref{tab:visual_budget}). Assigning 512 frames to past-routed queries and one frame to current-routed queries improves the overall score from 58.02 to 58.73, suggesting that the planner already provides a useful signal for adaptive visual budget allocation, while leaving more sophisticated policies to future work.

\subsection{Adaptive selection vs. fixed top-$K$.} Table~\ref{tab:ablation_selection} compares our online probabilistic lookback with a fixed top-$K$ retrieval variant under the same routing policy. We set $K=4$, which gives the fixed-retrieval baseline a slightly larger budget than the average number of entries selected by our online probabilistic method ($\bar{K}^{\star}{=}2.68$ over past-routed queries; Table~\ref{tab:kstar}). Replacing the adaptive cumulative-mass selection rule with this fixed top-$K$ strategy reduces backward accuracy from 51.37 to 49.64. This result shows that the number of relevant past entries should not be treated as constant across queries: some questions are resolved by a small amount of concentrated evidence, whereas others require a broader set of supporting captions. Our probabilistic selection mechanism adapts to this variability, preserving compactness when confidence is concentrated while expanding the evidence set when the retrieval distribution is more diffuse.

\begin{table}[t]
\centering
\caption{\textbf{Effect of adaptive past-evidence selection on OVO Backward Tracing.} Since this comparison isolates the past-lookback module, we report overall backward accuracy only. Results are obtained by applying our method on top of Qwen3-VL-8B\citep{Qwen3-VL}. We compare the proposed online probabilistic lookback with a fixed top-$K$ variant under the same routing setup. Replacing adaptive selection with fixed top-$K$ retrieval degrades performance.}
\vspace{2mm}
\scalebox{0.9}{
\begin{tabular}{@{}l c@{}}
\toprule
\textbf{Selection Strategy} & \textbf{Backward Avg.} \\
\midrule
Top-$K$ Selection ($K=4$) & 49.64 \\
\textbf{Online Probabilistic Selection (Ours)} & \textbf{51.37} \\
\bottomrule
\end{tabular}
}
\label{tab:ablation_selection}
\end{table}

\subsection{Number of retrieved captions per question type.}
\label{app:kstar}
\begin{table*}[t]
\centering
\caption{\textbf{Number of captions selected by past lookback ($K^{\star}$) across OVO-Bench question types.} Statistics are computed over the 1,473 query instances routed to past.}
\label{tab:kstar}
\vspace{2mm}
\resizebox{\textwidth}{!}{
\begin{tabular}{@{}l cccccc ccc ccc@{}}
\toprule
& \multicolumn{6}{c}{Real-Time Visual Perception} & \multicolumn{3}{c}{Backward Tracing} & \multicolumn{3}{c}{Forward Active Responding} \\
\cmidrule(lr){2-7} \cmidrule(lr){8-10} \cmidrule(lr){11-13}
& OCR & ACR & ATR & STU & FPD & OJR & EPM & ASI & HLD & REC & SSR & CRR \\
\midrule
Mean $K^{\star}$ & 2.09 & 2.27 & 2.24 & 2.20 & 1.43 & 2.61 & 2.54 & 2.80 & 2.52 & 2.77 & 3.21 & 2.68 \\
Median $K^{\star}$ & 2 & 2 & 2 & 2 & 1 & 2 & 2 & 2 & 2 & 2 & 3 & 2 \\
Max $K^{\star}$ & 7 & 6 & 6 & 7 & 2 & 5 & 7 & 10 & 9 & 13 & 10 & 19 \\
\bottomrule
\end{tabular}
}
\end{table*}

Table~\ref{tab:kstar} reports the number of captions $K^{\star}$ selected by past lookback for the 1,473 OVO-Bench query instances routed to past. Across the 12 question types, the mean $K^{\star}$ ranges from 1.43 to 3.21 and the median from 1 to 3. Overall, past retrieval selects 2.68 captions on average, corresponding to approximately 480 caption tokens under the accounting of Table~\ref{tab:text_vs_visual}, or about 0.18\% of the answer model's 256K-token context window. Only 10 of the 1,473 instances (0.68\%) have $K^{\star}>10$, all in REC and CRR; the largest case ($K^{\star}=19$) corresponds to about 3.4K caption tokens, or 1.3\% of the context window. The trend follows the temporal demands of each category: Real-Time instances retrieve the fewest captions on average (2.19), Backward instances more (2.61), and Forward instances the most (2.82), consistent with sequential-step and clue-based questions that need evidence accumulated across several moments. The retrieved context therefore remains small for all question types.

\subsection{Redundancy among retrieved captions.}
\label{app:dedup}
Using temporal proximity as a proxy for local redundancy, 28.2\% of adjacent selected-caption pairs lie within 1\,s of each other and 49.4\% within 2\,s. To test whether these nearby captions are redundant, we insert a greedy temporal-deduplication step after causal filtering and before top-$N$ truncation: candidates are processed in descending retrieval-score order, and a candidate is removed if it lies within 1\,s or 2\,s of a higher-scoring retained candidate, with all other components unchanged. On OVO Backward Tracing, a 1-s window reduces the mean $K^{\star}$ by 17\% and the text budget by 15\% without degrading accuracy (task-averaged $+$0.12, instance-averaged unchanged), whereas a 2-s window reduces them by 29\% and 27\% but lowers accuracy (task-averaged $-$0.44, instance-averaged $-$0.63). Local overlap is therefore present, but temporally nearby captions are not always redundant. The degree of redundancy likely depends on video content, such as the presence of shot boundaries and the balance between static dialog scenes and action.

\begin{table*}[t]
\centering

\begin{minipage}[t]{0.32\textwidth}
\centering
\refstepcounter{table}\label{tab:sensitivity_T}
\footnotesize Table \thetable. \textbf{Sensitivity to temperature $T_{\mathrm{past}}$ on OVO Backward Tracing.}\par
\vspace{0.5em}
\scalebox{0.96}{
\begin{tabular}{@{}c c@{}}
\toprule
\textbf{$T_{\mathrm{past}}$} & \textbf{Backward Avg.} \\
\midrule
0.01 & 50.39 \\
\textbf{0.02} & \textbf{51.37} \\
0.05 & 50.34 \\
0.10 & 50.16 \\
\bottomrule
\end{tabular}
}
\end{minipage}\hfill
\begin{minipage}[t]{0.32\textwidth}
\centering
\refstepcounter{table}\label{tab:sensitivity_tp}
\footnotesize Table \thetable. \textbf{Sensitivity to threshold $t_p$ on OVO Backward Tracing.}\par
\vspace{0.5em}
\scalebox{0.96}{
\begin{tabular}{@{}c c@{}}
\toprule
\textbf{$t_p$} & \textbf{Backward Avg.} \\
\midrule
0.10 & 50.72 \\
0.30 & 50.50 \\
0.50 & 50.84 \\
\textbf{0.70} & \textbf{51.37} \\
0.90 & 50.74 \\
1.00 & 50.07 \\
\bottomrule
\end{tabular}
}
\end{minipage}\hfill
\begin{minipage}[t]{0.32\textwidth}
\centering
\refstepcounter{table}\label{tab:sensitivity_tau}
\footnotesize Table \thetable. \textbf{Sensitivity to recency scale $\tau_r$ on OVO Backward Tracing.}\par
\vspace{0.5em}
\scalebox{0.96}{
\begin{tabular}{@{}c c@{}}
\toprule
\textbf{$\tau_r$} & \textbf{Backward Avg.} \\
\midrule
No recency bias & 49.51 \\
\textbf{150} & 51.37 \\
300 & 51.35 \\
600 & \textbf{51.51} \\
1200 & 50.87 \\
\bottomrule
\end{tabular}
}
\end{minipage}

\end{table*}

\begin{table}[t]
\centering
\caption{\textbf{Sensitivity to $T_{\mathrm{past}}$ across answer backbones on OVO-Bench Backward Tracing (average accuracy).} All other settings are fixed. The no-memory baseline is 43.44 for Qwen3-VL-8B and 32.25 for InternVL3.5-8B.}
\label{tab:temp_robust}
\vspace{2mm}
{\small
\begin{tabular}{@{}l cccc c@{}}
\toprule
\textbf{Answer model} & 0.01 & \textbf{0.02 (Ours)} & 0.05 & 0.10 & Range \\
\midrule
Qwen3-VL-8B & 50.39 & \textbf{51.37} & 50.34 & 50.16 & 1.21 \\
InternVL3.5-8B & 34.43 & 34.72 & \textbf{35.57} & 34.47 & 1.14 \\
\bottomrule
\end{tabular}
}
\end{table}

\subsection{Sensitivity to temperature $T_{\mathrm{past}}$.} Table~\ref{tab:sensitivity_T} examines the temperature $T_{\mathrm{past}}$ used in the recency-aware softmax for past-evidence selection. Since $T_{\mathrm{past}}$ is a hyperparameter of the past-lookback module, Table~\ref{tab:sensitivity_T} reports overall backward accuracy only. Results are obtained by applying our method on top of Qwen3-VL. The best backward-tracing accuracy is obtained at $T_{\mathrm{past}}=0.02$, which reaches 51.37. When the temperature is reduced to 0.01, performance drops to 50.39, suggesting that overly sharp distributions can over-commit to a very small set of candidates and become brittle to retrieval noise. Increasing the temperature to 0.05 or 0.10 also hurts performance, indicating that excessively smooth selection weakens the model's ability to focus on the most relevant historical evidence. These results support our choice of a moderately sharp temperature that balances selectivity and robustness. We further repeat the sweep with InternVL3.5 (Table~\ref{tab:temp_robust}). The Backward average varies by at most 1.21 points with Qwen3-VL and 1.14 points with InternVL3.5, and every setting improves over the corresponding no-memory baseline (by at least 6.72 and 2.18 points, respectively). The best value for InternVL3.5 is 0.05, which exceeds the adopted value by 0.85 points, indicating that the adopted setting transfers across backbones.

\subsection{Sensitivity to nucleus threshold $t_p$.} Table~\ref{tab:sensitivity_tp} studies the nucleus threshold $t_p$, which controls how much cumulative probability mass is retained in the adaptive evidence prefix. Since $t_p$ is a hyperparameter of the past-lookback module, Table~\ref{tab:sensitivity_tp} reports overall backward accuracy only. Results are obtained by applying our method on top of Qwen3-VL. The strongest result again appears at $t_p=0.70$, yielding 51.37 backward accuracy. Smaller thresholds such as 0.10 or 0.30 underperform, which suggests that overly aggressive truncation discards useful complementary context. On the other hand, very large thresholds such as 0.90 or 1.00 also degrade performance, implying that including too many lower-confidence entries introduces distracting historical evidence. This trend confirms that adaptive selection is most effective when it retains a compact but sufficiently informative subset of past captions.

\subsection{Sensitivity to recency scale $\tau_r$.} Table~\ref{tab:sensitivity_tau} evaluates the recency scale $\tau_r$, which determines how strongly older evidence is penalized during lookback. Since $\tau_r$ is a hyperparameter of the past-lookback module, Table~\ref{tab:sensitivity_tau} reports overall backward accuracy only. Results are obtained by applying our method on top of Qwen3-VL. Moderate recency scales perform best: $\tau_r$ between 150 and 600 yields 51.35--51.51 backward accuracy, while removing recency bias entirely (implemented by dropping the second term in Eq.~(\ref{eq:past_logit})) leads to a clear drop to 49.51. This gap shows that semantic similarity alone is insufficient for streaming QA, because temporally distant but semantically related captions can act as distractors. Performance declines again at 1200, where the temporal penalty becomes too weak. In particular, the competitive result at $\tau_r=600$ suggests that retrieval quality remains strong even over roughly 10 minutes of context. Overall, the results validate recency-aware retrieval as an important ingredient for temporally grounded evidence acquisition.

\begin{figure}[t]
\centering
\includegraphics[width=0.85\linewidth]{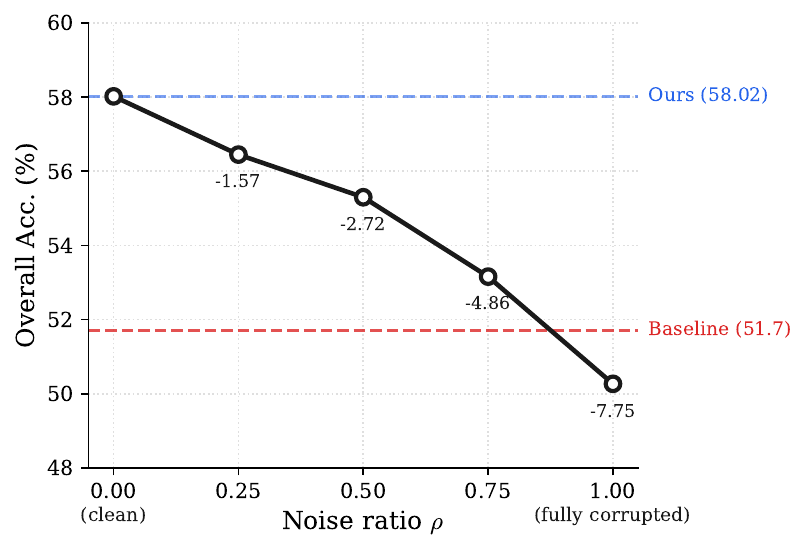}
\caption{\textbf{Textual input sensitivity on OVO-Bench.} We replace a fraction $\rho$ of the Time Dictionary captions with irrelevant captions sampled from unrelated videos and measure the resulting Overall accuracy. Dashed lines indicate the baseline (no memory) and our full method (clean captions).}
\label{fig:noise_injection}
\end{figure}
\subsection{Textual Input Sensitivity}
To verify that the Time Dictionary provides substantive evidence rather than merely additional context, we conduct a noise injection experiment (Fig.~\ref{fig:noise_injection}). We replace a fraction $\rho$ of the stored captions with irrelevant captions sampled from unrelated videos and evaluate on OVO-Bench. Performance degrades steadily as $\rho$ increases, dropping from 58.02 at $\rho{=}0$ to 50.27 at $\rho{=}1.0$, which falls below the no-memory baseline (51.70). This shows that the answer model actively relies on the retrieved caption content to produce correct answers, rather than benefiting from the mere presence of additional text input. It also validates the quality of our Time Dictionary: the captions generated by the streaming captioner carry accurate, answer-relevant evidence that meaningfully contributes to question answering.

\begin{figure}[t]
\centering
\includegraphics[width=\linewidth]{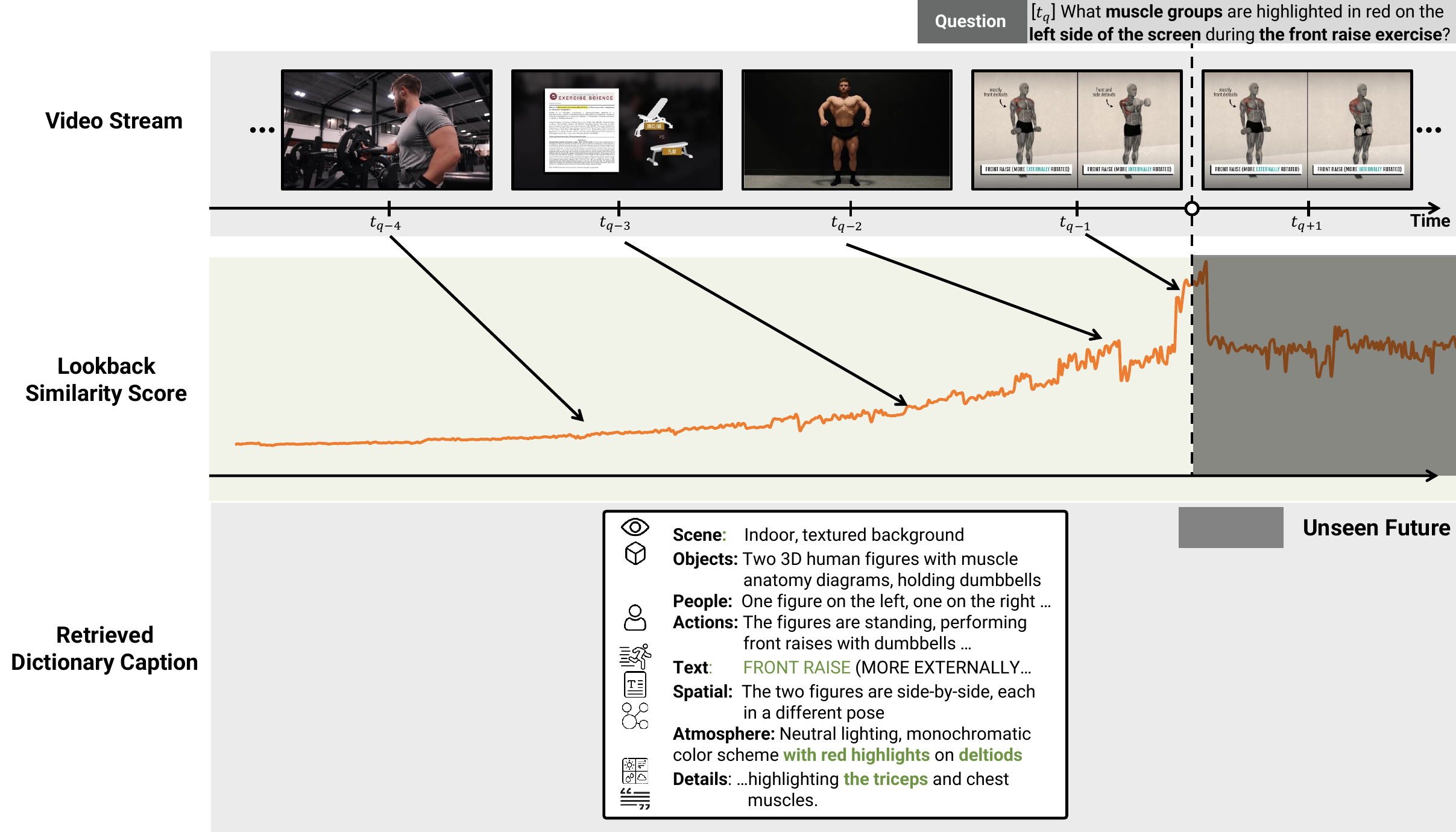}
\caption{Qualitative example of lookback retrieval on OVO-Bench. Given a query at time $t_q$, the similarity score over past captions (orange curve) peaks at the frame most relevant to the question. The retrieved Time Dictionary caption preserves fine-grained details such as on-screen text (\textit{``FRONT RAISE''}), spatial layout, and highlighted muscle groups, which are essential for answering correctly.}
\label{fig:qualitative_retrieval}
\end{figure}
\subsection{Qualitative Result of Lookback and Dictionary.}
Fig.~\ref{fig:qualitative_retrieval} illustrates a representative lookback retrieval case. The user asks which muscle groups are highlighted during a front raise exercise. The similarity score peaks at a past frame showing a 3D anatomy diagram with red-highlighted deltoids, and the corresponding Time Dictionary caption retains the precise on-screen text, object descriptions, and spatial relations needed to answer. This example demonstrates that our structured captions preserve the fine-grained evidence that compressed visual representations would likely discard, and that the retrieval mechanism successfully surfaces the most relevant past observation.

\clearpage
\section{Prompts in \methodname}
\label{sec:prompts}

We document the three prompt templates used in our pipeline:
the per-frame \textbf{captioner} that populates the Time
Dictionary (\S\ref{sec:prompts:captioner}), the \textbf{planner}
that performs temporal routing
(\S\ref{sec:prompts:planner}), and the \textbf{gating} model that
decides cue-trigger emission for proactive-output questions
(\S\ref{sec:prompts:gating}).

\subsection{Captioner Prompt}
\label{sec:prompts:captioner}

Per-frame captions populating the Time Dictionary are produced
by a lightweight VLM (8-bit-quantized Qwen3-VL-2B~\citep{Qwen3-VL}) using the
structured 8-field template below. The prompt enforces concision, non-redundancy across fields, and a fixed field order so that downstream schema-level operations (e.g., field-routed retrieval, schema-drop ablation) can address each field deterministically.
Fields with no observable content are filled with the literal string ``N/A''.

\begin{tcolorbox}[
  colback=gray!5, colframe=gray!60, boxrule=0.5pt,
  title=\textbf{Captioner Prompt}, fonttitle=\bfseries,
  left=4pt, right=4pt, top=4pt, bottom=4pt,
  breakable
]
\ttfamily\small
Analyze the image and provide a structured description using the
format below. Be concise. List key items only, do not repeat
information across fields. Focus on the most salient elements. If
a field is not applicable, write `N/A'.\\[3pt]
Scene: [locational or environmental information,
indoor/outdoor]\\
Objects: [key objects with attributes such as color, size,
state]\\
People: [count, appearance, clothing, expression]\\
Actions: [what is happening, interactions between
people/objects]\\
Text: [any visible text, signs, labels, subtitles, logos]\\
Spatial: [layout and relative positions of key elements]\\
Atmosphere: [lighting, color tone, time of day, weather,
mood]\\
Details: [description sentences that are shown in the
image in one coherent paragraph. Write 2--3 sentences and keep it
concise, around 80--120 words.]
\end{tcolorbox}

\noindent\textbf{Decoding configuration.}
Greedy decoding with $\text{temperature}=0$, $\text{top-}p=0.8$,
$\text{top-}k=-1$, presence penalty $=1.5$, repetition penalty
$=1.0$, and max new tokens $=400$.

\subsection{Planner Prompt}
\label{sec:prompts:planner}

Each incoming question is routed into one of three temporal
regions (\texttt{past}, \texttt{current}, or \texttt{future}) by a
caption-augmented meta-classifier (8-bit-quantized Qwen3-VL-2B, text-only inference). In addition to the question, the planner receives the
\emph{current frame caption} -- the most-recent caption from the
Time Dictionary -- so that routing decisions are grounded in
present visual context. The decision selects the retrieval scope
used downstream: specific past captions retrieved by query
similarity (\texttt{past}), the most-recent frame caption from
the temporal buffer (\texttt{current}), or a wait-and-emit gating
loop (\texttt{future}).

\begin{tcolorbox}[
  colback=gray!5, colframe=gray!60, boxrule=0.5pt,
  title=\textbf{Planner Prompt (curcap v11)}, fonttitle=\bfseries,
  left=4pt, right=4pt, top=4pt, bottom=4pt,
  breakable
]
\ttfamily\small
You route streaming video questions into one of three temporal
regions based on what the answer depends on.\\[3pt]
\textbullet\ \textbf{current}: the answer is a property of the
frame at the moment of asking. It does not require looking back
or aggregating earlier content.\\[2pt]
\textbullet\ \textbf{past}: the answer requires information from
earlier in the video. This includes counting things that have
occurred, summarizing prior activity, comparing with an earlier
moment, reasoning about causes or sequences, or referring to a
specific earlier event, object, or statement.\\[2pt]
\textbullet\ \textbf{future}: the question sets up a trigger ---
the system must wait and emit a response only when a specified
later condition occurs. Typical cues: ``when X happens, output
Y'', instructions to monitor and respond.\\[5pt]
Present-tense phrasing alone does not imply current. If the
correct answer needs evidence from before this moment, it is
past.\\[5pt]
Current frame caption: ``\{\textit{caption}\}''\\
Question: ``\{\textit{question}\}''\\[3pt]
Reply with exactly one word: current, past, or future.
\end{tcolorbox}

\noindent\textbf{Decoding configuration.}
Greedy decoding ($\text{temperature}=0$,
$\text{do\_sample}=\text{False}$), max new tokens $=8$. The
single-word output is parsed case-insensitively; malformed
outputs default to \texttt{current}.

\noindent\textbf{Caption substitution.}
The \texttt{\{caption\}} placeholder is filled with the
8-field structured caption (\S\ref{sec:prompts:captioner}) of the
frame nearest to the question's time stamp.

\subsection{Gating Prompt}
\label{sec:prompts:gating}
For proactive-output questions (the question sets up a future
trigger of the form ``when X happens, output Y''), a binary
gating prompt is issued at each streaming step to decide whether
the cue condition has been satisfied at the current moment. When
the gate emits ``yes'', a follow-up cue-response call is issued
on the same clip to extract the actual cue answer; the
$(\text{time}, \text{answer})$ tuple is then scored by the evaluator against the
ground-truth event time and the specified output string.

\noindent\textbf{Source of \texttt{\{specified\_output\}}.}
The placeholder \texttt{\{specified\_output\}} holds the response text $Y$
that the user instruction itself asks the system to emit (``\ldots, output $Y$'').
Following the official StreamingBench protocol~\citep{lin2026streamingbench},
this text is parsed from the Proactive Output instruction and is part of the
standard evaluation input, equally available to all evaluated methods.
The ground-truth trigger timestamp is used only by the evaluator.
At each step, the gate decides whether to respond or wait using only the
question, the specified output, and captions obtained causally up to the
current time.

\begin{tcolorbox}[
  colback=gray!5, colframe=gray!60, boxrule=0.5pt,
  title=\textbf{Gating Prompt},
  fonttitle=\bfseries,
  left=4pt, right=4pt, top=4pt, bottom=4pt,
  breakable
]
\ttfamily\small
You are a gating agent watching a streaming video at time t=\{\textit{t\_now}\}s.\\
At t=\{\textit{t\_q}\}s the user asked: ``\{\textit{question}\}''\\
The expected output to deliver when the right moment arrives: ``\{\textit{specified\_output}\}''\\
Captions describing what is visible:\\
{[}t=\{\textit{t\_now}\}s{]} \{\textit{caption}\}\\
Based on the captions, has the moment to deliver the expected output arrived? Answer with only ``yes'' or ``no'':
\end{tcolorbox}

\noindent\textbf{Decoding configuration.}
Greedy decoding ($\text{temperature}=0$,
$\text{do\_sample}=\text{False}$), max new tokens $=32$. The
gate output is parsed by case-insensitive substring containment
of ``yes''. The cue-response output is scored by case-insensitive
substring containment of \texttt{specified\_output}.

\noindent\textbf{Caption substitution.}
The \texttt{\{caption\}} placeholder is filled with the
8-field structured caption (\S\ref{sec:prompts:captioner}) of the
frame at the current polling time \texttt{t\_now} (window size $W{=}1$).

\section{Additional Benchmark Explanation}

\subsection{OVO-Bench.}
OVO-Bench~\citep{niu2025ovo} evaluates temporal awareness in online video understanding by posing questions at specific timestamps during a video stream, rather than after the entire video has been observed.
The benchmark comprises 644 videos with 2,814 human-curated question-answer pairs across 12 tasks, organized into three temporal scenarios.
\textit{Real-Time Visual Perception} tests understanding of the current scene at query time, covering Optical Character Recognition (OCR), Attribute Recognition (ACR), Action Recognition (ATR), Spatial Understanding (STU), Future Prediction (FPD), and Object Recognition (OJR).
\textit{Backward Tracing} requires the model to recall past events, with tasks including Episodic Memory (EPM), Action Sequence Identification (ASI), and Hallucination Detection (HLD).
\textit{Forward Active Responding} presents questions whose answers depend on events that have not yet occurred; the model must decide to wait rather than respond prematurely. This scenario includes Repetition Event Count (REC), Sequential Steps Recognition (SSR), and Clues Reveal Responding (CRR).
Backward Tracing and Real-Time Visual Perception adopt a multiple-choice format with accuracy as the metric.
Forward Active Responding employs a multiple-triggering evaluation pipeline that densely queries models along the temporal axis, using both accuracy-based and score-based metrics.
The overall score is the macro average across the three scenario-level averages.

\noindent\textbf{Evaluation protocol for Forward Active Responding.}\label{app:ovo_protocol} OVO-Bench Forward Active Responding follows the official multiple-triggering protocol~\citep{niu2025ovo}: models are densely queried along the timeline, each round receives the clip from 0\,s up to that polling time, and rounds are scored independently. We follow this protocol identically for all methods, so the deferred-observation gate (Eq.~(\ref{eq:future_gate})) does not intervene on OVO Forward, and the planner routes its 172 questions to past (104), current (67), and future (1). The future gate is instead evaluated on StreamingBench Proactive Output, where the system itself must decide when to respond.

\subsection{StreamingBench.}
StreamingBench~\citep{lin2026streamingbench} evaluates streaming comprehension by presenting questions at different temporal positions within a video, simulating real-time user interaction during playback.
The benchmark contains 900 videos with 4,500 human-curated QA pairs across 18 tasks under three dimensions.
\textit{Real-Time Visual Understanding} (10 tasks) covers Object Perception (OP), Causal Reasoning (CR), Clips Summarization (CS), Attribute Perception (ATP), Event Understanding (EU), Text-Rich Understanding (TR), Prospective Reasoning (PR), Spatial Understanding (SU), Action Perception (ACP), and Counting (CT).
\textit{Omni-Source Understanding} (4 tasks) requires integrating audio and visual signals, including Emotion Recognition (ER), Scene Understanding (SCU), Source Discrimination (SD), and Multimodal Alignment (MA).
\textit{Contextual Understanding} (4 tasks) evaluates higher-level reasoning over accumulated context, including Anomaly Context Understanding (ACU), Misleading Context Understanding (MCU), Sequential Question Answering (SQA), and Proactive Output (PO).
The PO task is notable in that the model must determine the appropriate moment to respond without an explicit user query; a response is correct only if the output timestamp falls within two seconds of the ground-truth timestamp.
All tasks follow a multiple-choice format with accuracy as the primary metric.

\section{Limitations and Broader Impacts}
\label{app:limitations}
\paragraph{Limitations and future work.} The planner outputs a single temporal strategy per query and therefore does not explicitly model implicit future intent or queries that require combining past, current, and future evidence; confidence-aware or multi-route planning is a natural extension. The future gate responds at its first positive decision, so instructions that require accumulating repeated events over a fixed future interval (e.g., counting events during the next ten seconds) would require maintaining and updating an accumulated count until the horizon, a setting not covered by current benchmarks. Our system is visual-only, so audio-dependent tasks remain outside its scope. Broader validation under noisier real-world streams, multi-turn interactions, and longer future waiting horizons remains future work. Finally, since structured textual memory is complementary to visual memory methods such as KV-cache compression~\citep{kim2025infinipot}, combining the two is another promising direction.

\paragraph{Broader impacts.}
Our work improves streaming video question answering by preserving and retrieving temporal evidence from video streams. Potential negative impacts include privacy-sensitive deployment in surveillance, workplace monitoring, or bystander recording, where persistent textual memory may expose identities, locations, or sensitive activities. Incorrect captions or temporal misrouting may also cause false alarms or misleading summaries in safety-critical settings. We recommend deployment only with user notification, access control, retention limits for the Time Dictionary, and mechanisms for deleting stored temporal records.

\end{document}